\documentclass{article}

\usepackage{microtype}
\usepackage{graphicx}
\usepackage{subcaption}
\usepackage{booktabs}

\usepackage{hyperref}

\usepackage[accepted]{preprintcustom}

\icmltitlerunning{Analyzing Neural Networks Based on Random Graphs}

\usepackage{amsmath}
\usepackage{xspace}

\newcommand{\eq}{\begin{equation}}
\newcommand{\eqx}{\end{equation}}

\newcommand{\ba}{{\tt ba}\xspace}
\newcommand{\er}{{\tt er}\xspace}
\newcommand{\ws}{{\tt ws}\xspace}
\newcommand{\rdag}{{\tt rdag}\xspace}
\newcommand{\rrdags}{{\tt rdags}\xspace}
\newcommand{\fmri}{{\tt fmri}\xspace}
\newcommand{\bottleneck}{{\tt bottleneck}\xspace}
\newcommand{\composite}{{\tt composite}\xspace}

\newcommand{\bifocal}{{\tt bf}\xspace}
\newcommand{\radial}{{\tt r}\xspace}
\newcommand{\radialr}{{\tt rr}\xspace}
\newcommand{\x}{{\tt x}\xspace}

\newcommand{\python}{{\tt python}\xspace}
\newcommand{\networkx}{{\tt networkx}\xspace}
\newcommand{\slearn}{{\tt scikit-learn}\xspace}

\newcommand{\pytorch}{{\tt pytorch}\xspace}

\newcommand{\f}[2]{\frac{#1}{#2}}

\begin{document}

\twocolumn[
\icmltitle{Analyzing Neural Networks Based on Random Graphs}

\begin{icmlauthorlist}
\icmlauthor{Romuald A. Janik}{ujf}
\icmlauthor{Aleksandra Nowak}{ujm}
\end{icmlauthorlist}

\icmlaffiliation{ujf}{Jagiellonian University, Institute of Theoretical Physics,  \L ojasiewicza~11, 30-348~Kraków,~Poland.}
\icmlaffiliation{ujm}{Jagiellonian University, Faculty of Mathematics and Computer Science, \L ojasiewicza~6, 30-348~Kraków,~Poland}

\icmlcorrespondingauthor{Romuald A. Janik}{romuald.janik@gmail.com}
\icmlcorrespondingauthor{Aleksandra Nowak}{nowak.aleksandrairena@gmail.com}

\vskip 0.3in
]

\printAffiliationsAndNotice{} 

\begin{abstract}
We perform a massive evaluation of neural networks with architectures corresponding to random graphs of various types. 
We investigate various structural and numerical properties of the graphs in relation to neural network test accuracy.
We find that none of the classical numerical graph invariants by itself allows to single out the best networks. Consequently, we introduce a new numerical graph characteristic that selects a set of \emph{quasi-1-dimensional} graphs, which are a majority among the best performing networks.
We also find that networks with primarily short-range connections perform better than networks which allow for many long-range connections.
Moreover, many resolution reducing pathways are beneficial. 
We provide a dataset of 1020 graphs and the test accuracies of their corresponding neural networks at
\url{https://github.com/rmldj/random-graph-nn-paper}



\end{abstract}

\section{Introduction}
\label{s.introduction}
The main aim of this paper is to perform a wide ranging study of neural networks based on a variety of random graphs and analyze the interrelation of the structure of the graph with the performance of the corresponding neural network. The motivation for this study is twofold. 

On the one hand, artificial neural networks typically have a quite rigid connectivity structure. Yet in recent years significant advances in performance have been made through novel global architectural changes like ResNets, \cite{he2016deep_residual} or DenseNets \cite{huang2017densely}. This has been further systematically exploited in the field of Neural Architecture Search (NAS, see \cite{elsken2019nas_survey} for a review).  Hence there is a definite interest in exploring a wide variety of possible global network structures. 
On the other hand, biological neural networks in the brain do not have a rigid structure and some randomness is an inherent feature of networks which evolved ontogenetically.
Contrarily, we also do not expect these networks to be totally random. Therefore, it is very interesting to investigate the interrelations of structural randomness and global architectural properties with the network performance.

To this end, we explore a wide variety of neural network architectures constructed accordingly to wiring topologies defined by random graphs. This approach can efficiently produce many qualitatively different connectivity patterns by alternating only the random graph generators. 

The nodes in the graph correspond to a simple computational unit, whose internal structure is kept fixed. Apart from that, we do not impose any restrictions on the overall structure of the neural network. In particular, the employed constructions allow for modelling 
arbitrary global (as well as local) connectivity.

We investigate a very wide variety of graph architectures, which range from the quintessential random, scale-free and small world families, through some novel algorithmic constructions, to graphs based on  fMRI data. Altogether we conduct an analysis of more than 1000 neural networks, each corresponding to a different directed acyclic graph.
Such a wide variety of graphs is crucial for our goal of analyzing the properties of the network architecture by studying various characteristics of the corresponding graph and examining their impact on the performance of the model.

The paper is organized as follows. In section~\ref{s.relatedwork}, we discuss the relation to previous work and describe, in this context, our contribution. In section~\ref{s.nndetails}, we summarize the construction of the neural network architecture associated with a given directed acyclic graph. In section~\ref{s.space}, we discuss in more detail the considered space of graphs, focusing on the new families.
Section~\ref{s.keyresults} contains our key results, including the identification of the best and worst networks and the introduction of a novel numerical characteristic which enables to pick out the majority of the best performing graphs. We continue the analysis in section~\ref{s.architecture}, where 
we analyze the impact on network performance of various architectural features like resolution changing pathways, short- vs. long-range connectivity and depth vs. width.
We close the paper with a summary and outlook.

\section{Related Work}
\label{s.relatedwork}

\paragraph{Neural Architecture Search.}
Studies undertaken over the recent years indicate a strong connection between the wiring of network layers and its generalization performance. For instance, ResNet introduced by \cite{he2016deep_residual}, or DenseNet proposed in \cite{huang2017densely}, enabled successful training of very large multi-layer networks, only by adding new connections between regular blocks of convolutional operations. The possible performance enhancement that can be gained by the change of network architecture has posed the question, whether the process of discovering the optimal neural network topology can be automatized. In consequence, many approaches to this Neural Architecture Search (NAS) problem were introduced over the recent years \cite{elsken2019nas_survey}. Among others, algorithms based on reinforcement learning \cite{zoph2016neural, baker2016designing}, evolutionary techniques \cite{real2019regularizedAmoebaNet} or differentiable methods \cite{liu2018darts}. Large benchmarking datasets of the cell-operation blocks produced in NAS  have been also proposed by \cite{ying2019bench, dong2019bench}.

\paragraph{Differences with NAS.} There are two key differences between the present work and the investigations in NAS. Firstly, the NAS approaches focus predominantly on optimizing a rather intricate structure of local cells which are then combined into a deep network with a relatively simple linear global pattern (e.g. \cite{ying2019bench, real2019regularizedAmoebaNet}). The main interest of the present paper is, in contrast, to allow complete flexibility \emph{both} in the local and global structure of the network (including connections crossing all resolution stages), while keeping the architecture of the elementary node fixed. Secondly, we are not concentrating on directly optimizing the architecture of a neural network for performance, but rather on exploring a wide variety of random graph architectures in order to identify what features of a graph are related to good or bad performance of the associated neural network.
This goal necessitates an approach orthogonal to NAS in that we need to study both strong and weak architectures in order to ascertain whether a given feature is, or is not \emph{predictive} of good performance.

\paragraph{Random Network Connectivity.}

There were already some prior approaches which
focused on introducing randomness or irregularity into the network connectivity pattern. The work \cite{shafiee2016stochasticnet} proposed stochastic connections between consecutive feed-forward layers, while in \cite{huang2016stochastic_depth} entire blocks of layers were randomly dropped during training.

However, the first paper which, to our knowledge, really investigated neural networks on random geometries was the pioneering work of \cite{xie2019exploring}. This paper proposed a concrete construction of a neural network based on a set of underlying graphs (one for each resolution stage of the network). Several models based on classical random graph generators were evaluated on the ImageNet dataset, achieving competitive results to the models obtained by NAS or hand-engineered approaches. Using the same mapping, very recently \cite{roberts2019deepConnectomicsNetworks} investigated neural networks based on the connectomics of the mouse visual cortex and the biological neural network of \emph{C.Elegans}, obtaining high accuracies on the MNIST and FashionMNIST datasets. 

Although the works discussed above showed that deep learning models based on random or biologically inspired architectures can indeed be successfully trained without a loss in the predictive performance, they did not investigate what kind of graph properties characterize the best (and worst) performing topologies. 

The idea of analyzing the architecture of the network by investigating its graph structure has been raised in \cite{you2020graph}.  However, this work focused on exploring the properties of the introduced relational graph, which defined the communication pattern of a network layer. Such pattern was then repeated sequentially to form a deep model. 

\paragraph{Our Contribution.} The main goal of our work is to perform a detailed study of numerical graph characteristics in relation to the associated neural network performance. Contrary to \cite{you2020graph} we are not concentrating on exploring the fine-grained architecture of a layer in a sequential network. Instead, we keep the low-level operation pattern fixed (encapsulated in the elementary computational node) and focus on the higher level connectivity of the network, by analyzing the graph characteristics of neural network architectures based on arbitrary directed acyclic graphs (DAG)s. Our models are obtained by the use of a mapping similar to the one presented in \cite{xie2019exploring}. Apart from the quintessential classical families of Erd\H{o}s-R\'{e}nyi, small world and scale-free graphs used in that paper, we introduce a novel and flexible way of directly generating random DAGs and also investigate a set of graphs derived from functional fMRI networks from Human Connectome Project data \cite{HCPmain}. Altogether we performed a massive empirical study on CIFAR-10 of 1020 neural networks each corresponding to a different graph. We also evaluated 450 of these networks on CIFAR-100 in order to ascertain the consistency in the behaviour of various graph families.

\section{From a Graph to a Neural Network}
\label{s.nndetails}

In order to transform a graph to a neural network, we  essentially adopt the approach presented in \cite{xie2019exploring}. In that paper, a graph is sampled from a predefined list of generators and transformed to a DAG. Next, the DAG is mapped to a neural network architecture as follows:

The edges of the graph represent the flow of the information in the network and the nodes correspond to the operations performed on the data. For each node, the input from the ingoing edges is firstly aggregated using a weighted sum. Next, a ReLU – Conv2d – Batch-Norm block is applied. The result of this procedure is then propagated independently by each outgoing edge. The only node that does not follow this construction is the output node, which additionally performs a global average pooling on the weighted sum of its inputs and then applies a dense layer with the number of output neurons equal to the target dimension. Finally, the network nodes are divided into three sets, referred to as \emph{stages} (denoted by different colours in the figures). The first stage operates on the original input resolution, with the number of channels $C$ being set in the first (input) node of the graph. The subsequent stages operate on a decreased input resolution and increased  number of output channels by a factor of 2, with respect to the previous stage. In order to perform the downsampling, on every edge that crosses the stages  the same block of operations as in a standard node is executed, but with the use of convolutions with~stride~2. In the figures in the present paper, we represent such resolution changing edges with beige color. 

We introduce three modifications to this procedure:

Firstly, in \cite{xie2019exploring} there were separate random graphs for each stage of the neural network. In our case, we have a random graph for the \emph{whole} network and dimensionality reduction is performed on a graph edge when necessary, by a node with stride 2 or 4 convolutions and a single input path.
In consequence, we do not  bias the model to have single bottleneck connection between the computations performed on different spatial resolutions. Moreover, we observe that the introduction of such bottleneck generally deteriorates the network performance (we discuss this issue in section~\ref{s.bottleneck}).

\begin{figure}[ht]
\vskip 0.1in
\centerline{\includegraphics[width=0.35\textwidth]{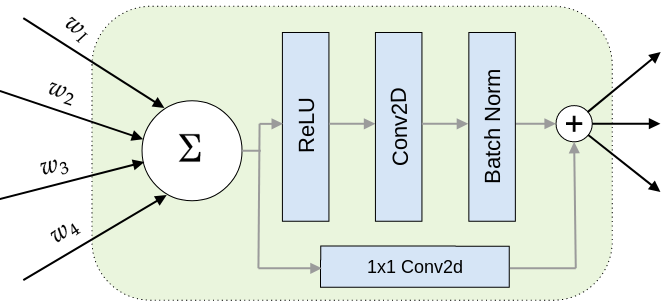}}
\caption{The node is represented by the green-shaded area. The black arrows illustrate the graph edges labeled with the associated weights. The gray arrows indicate the ordering of the operations performed in the node as well as the residual connection.}
\label{fig.node}
\vskip -0.1in
\end{figure}

Secondly, we introduce an additional residual connection from the aggregated signal to the output of the triplet block in the node. The residual connection always performs a projection (implemented by a $1$$\times$$1$ – convolution, similar to ResNet  C-type connections~\cite{he2016deep_residual} - see Figure \ref{fig.node}). The residual skip connection shifts the responsibility of taking care of the vanishing gradient problem from edges to the nodes, allowing the global connectivity structure to focus on the information flow, with the low-level benefits of the residual structure already built in. 

\begin{figure}[ht]
\begin{center}
\centerline{\includegraphics[height=2.8cm]{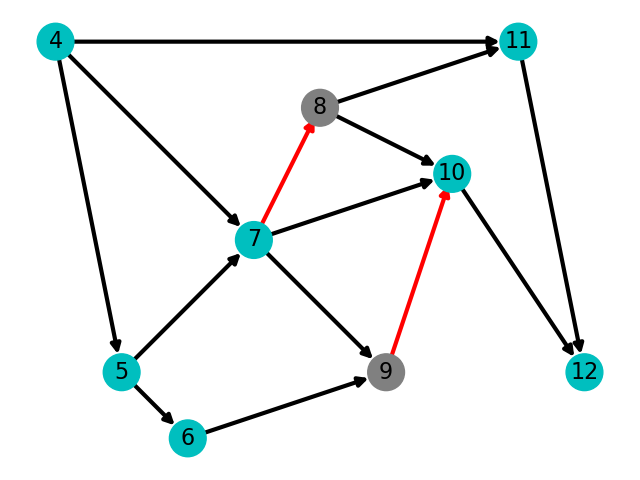}}
\caption{The gray nodes (orphan nodes) in the DAG either do not have an input from previous stages of processing or do not have an output. Hence we add the red edges from the immediately preceding node or to the immediately succeeding node.}
\label{fig.orphan}
\end{center}
\vskip -0.3in
\end{figure}

Thirdly, we improve the method of transforming a graph into a DAG so that it automatically takes into account the graph structure. This is achieved by ordering the nodes accordingly to a 2D Kamada-Kawai embedding \cite{kamada1989algorithm} and setting the directionality of an edge from the lower to the higher node number. Any arising orphan nodes like the ones in Figure \ref{fig.orphan} are then fixed by adding a connection from the node with the preceding number or adding a connection to the node with the succeeding number. We observe that this approach leads to approximately $2$x less orphan nodes than the random ordering, and circa $1.5$x less than the original ordering returned by the generator, which was used in \cite{xie2019exploring}. A detailed description of the DAG transformation process together with a comparison of various node orderings can be found in \textbf{Supplementary Materials~B and C}.



\begin{figure}[ht]
\centerline{
\includegraphics[width=0.42\textwidth]{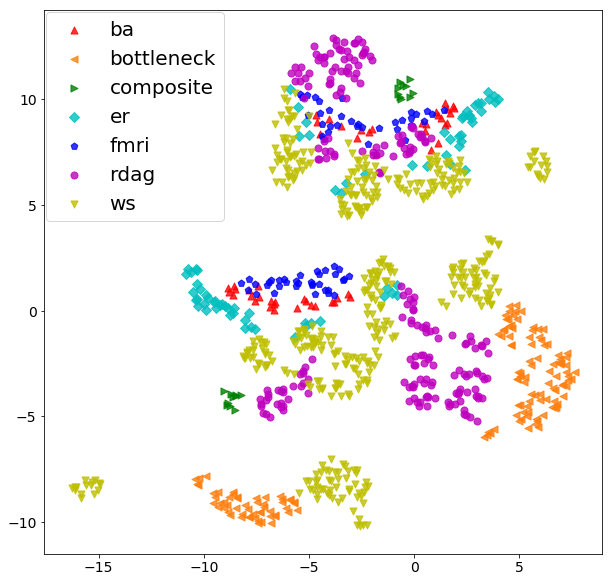}}
\caption{The UMAP embedding of the space of neural networks analyzed in the present paper, obtained from the dataset of the corresponding graph features  mentioned in section~\ref{s.graphinv}. Different colors represent different graph families. The central blob includes graphs with $n=60$ nodes.}
\label{fig.umap}
\vskip -0.2in
\end{figure}

\section{The Space of Random Graphs and DAGs}
\label{s.space}

We performed a massive empirical study of over $1000$ neural network architectures based on $5$ graph families and $2$ auxiliary constructions (see Fig.~\ref{fig.umap}). We summarize below their main characteristics.

{\bf Erd\H{o}s-R\'{e}nyi} (\er) – In this model, given a parameter $p\in[0,1]$, each possible (undirected) edge arises independently of all the other edges with probability $p$ \cite{erd6s1960evolution}. 

{\bf Barab\'{a}si-Albert} (\ba) –  
The Barab\'{a}si-Albert model favors the formation of hubs, as the few nodes with high degree are more likely to get even more connections in each iteration. Therefore graphs produced by this model are associated with scale-free networks. Apart from the number of nodes these graphs have a single integer parameter \cite{barabasi1999emergence}.

{\bf Watts-Strogatz} (\ws) - The graphs obtained by this method tend to have the small-world property. There are two nontrivial parameters: an integer and a real probability \cite{watts1998collective}.

{\bf Random-DAG} (\rdag) - The models  mentioned so far produce undirected graphs, which need to be later transformed to DAGs. 
We propose a new algorithm that instead directly constructs a random DAG. An advantage of this algorithm over existing DAG-generating methods is that it allows to easily model neural networks with mostly short-range or mostly long-range connections, which was the main reason for introducing this construction. This procedure and its parameters is thoroughly explained in section~\ref{s.rdag}.

{\bf fMRI based} (\fmri) - In addition to the above algorithmic generators we also introduce a family of graphs that are based on resting state functional MRI data from the Human Connectome Project. The exact method used to derive DAGs from the fMRI partial correlation matrices is described in detail in section~\ref{s.fmri}. Apart from the number of nodes, this family has a single thresholding parameter.

Moreover, we considered two auxiliary types of graphs:

{\bf Bottleneck graphs} (\bottleneck) - For some graphs from the above families, we introduced a bottleneck between the various resolution stages (see section~\ref{s.bottleneck}).

{\bf Composite graphs} (\composite) - We obtained these graphs by maximizing in a Monte-Carlo simulation the expression\\ 
\begin{equation}
    \left(\frac{log\_num\_paths}{num\_nodes} \right)^{\frac{1}{2}} -2 grc -avg\_clustering
\end{equation}
where $grc$ is the global reaching centrality of the graph. 
This construction was motivated by a certain working hypothesis investigated at an early stage of this work, but nevertheless we kept the graphs for additional structural variety. 

For each of the above families we fix a set of representative parameters\footnote{Refer to {\bf Supplementary Materials G} for a full list.}. Then for every family-parameters pair we sample $5$ versions of the model by passing different random seeds to the generator. Using this procedure we create $475$ networks with 30 nodes and $545$ networks with 60 nodes. We train all networks for 100 epochs with the same settings on the CIFAR-10 dataset\footnote{We provide a full description of the training procedure in \textbf{Supplementary Materials A}.}. For each network we set the number of initial channels $C$ in order to obtain approximately the same number of parameters as in ResNet-56 (853k).

\begin{algorithm}[ht]
   \caption{Random DAG}
   \label{alg.rdag}
\begin{algorithmic}
   \STATE {\bfseries Input:} nodes $i=0,\ldots,N-1$,\\ 
   \hspace{1.0cm} number of outgoing edges $n_i^{out}$,\\
   \hspace{1.0cm} size of a local neighbourhood $B$,\\ 
   \hspace{1.0cm} real $\alpha$, function $f(x)$
   \FOR{$i=0$ {\bfseries to} $N-2$}
   \IF{node $i+1$ does not have an ingoing connection}
   \STATE make an edge $i \to i+1$
   \ENDIF
   \WHILE{not all $n_i^{out}$ outgoing edges chosen}
   \STATE Make randomly the edge $i \to j$ with probability
   \[
   p_j = \f{w_{ij}}{\sum_{j>i} w_{ij}}
   \]
   where the weight $w_{ij}$ is given by
   \[
   w_{ij} = (n_j^{out})^\alpha f(\lfloor \frac{j-i}{B} \rfloor)
   \]
   \STATE provided $j>i$ and $i\to j$ does not exist so far
   \ENDWHILE
   \ENDFOR
\end{algorithmic}
\end{algorithm}

\subsection{Direct Construction of Random DAGs}
\label{s.rdag}


In order to study some specific questions, like the role of long-range versus short-range connectivity, we introduce a procedure for directly constructing random DAGs which allows for more fine-grained control than the standard random graph generators and is flexible enough to generate various qualitatively different kinds of graph behaviours. As an additional benefit, we do not need to pass through the slightly artificial process of transforming an arbitrary undirected graph to DAG. 

We present our method in Algorithm~\ref{alg.rdag}. We start with $N$ nodes, with a prescribed ordering given by integers $0,\ldots,N-1$. For each node $i$, we fix the number of outgoing edges $n_i^{out}$ (clearly $n_i^{out}<N-i$). Here we have various choices leading to qualitatively different graphs. For example sampling $n_i^{out}$ from a Gaussian and rounding to a positive integer (or setting $n_i^{out}$ to a constant) would yield approximately homogeneous graphs. Taking a long tailed distribution would yield some outgoing hubs. One could also select the large outgoing hubs by hand and place them in a background of constant and small $n_i^{out}$.

For each node $i$ we then randomly choose (with weight $w_{ij}$ given in Algorithm~\ref{alg.rdag}) nodes $j>i$ to saturate the required $n_i^{out}$ connections. The freedom in the choice of weight $w_{ij}$ gives us the flexibility of preferential attachment (through the parameter $\alpha$) and/or imposing local/semi-local structure (through the choice of function $f(x)$).

Different choices of $f(x)$ lead to different connectivity structures of the DAG. An exponential $f(x) = \exp(-C x)$ leads to short-range connections and local connectivity. The power law scaling
$f(x) = 1/x$
leads to occasional longer range connections, while
$f(x) = 1$
does not lead to any nontrivial spatial structure at all. In this work, we investigated the above three possibilities.

Since we do not want the integer node labels $i$ or $j$ to be effectively a 1d coordinate, we define a local neighbourhood size $B$ so that differences of node labels of order $B$ would not matter.
This motivates the form of the argument of the weighting function $f(x) \equiv f\left(  \lfloor \frac{j-i}{B} \rfloor \right) $, 
where $\lfloor a \rfloor$ denotes the floor of $a$. In the simulations we set $B=5$ or $B=10$.

As the above algorithm has several moving parts, let us summarize their roles. Firstly, through the choice of the function $f(.)$, we can model graphs with varying proportion of short to long range connections with the parameter $B$ defining the size of the local neighbourhood.
The choice of multiplicity distribution of $n_i^{out}$ allows to model, within the same framework, a uniform graph, a graph with power law outgoing degree scaling or a graph with a few hubs with very high multiplicity. Finally, the parameter $\alpha$ enables to control preferential attachment of connections. Consequently, the algorithm allows to produce DAGs with diverse architectural characteristics well suited for neural network analysis.

\subsection{Graphs Associated with fMRI Networks}
\label{s.fmri}

In this paper we supplement the families of algorithmically generated random graphs by including a family of  graphs derived from resting state connectome from fMRI data. We use the network connectomes provided by the Human Connectome Project \cite{HCPmain} based on resting state fMRI data of 1003 subjects \cite{HCPfMRI}. As an input for graph construction, we used the released (z-score transformed) partial correlation matrix for 50- and 100-component spatial group-ICA parcellation. 


In order to obtain a 30- or 60- node graph, we take the absolute value of the entries of the partial correlation matrix (of the 50- and 100- component version respectively) and use a range of thresholds\footnote{From $2.0$ to $5.0$ (or $4.9$ for the 50-component case) in steps of size $0.5$.} to binarize the matrix. Such matrix is then interpreted as an adjacency matrix of a graph. Since \emph{a priori} the graph obtained in this way does not need to be connected, we take the largest connected component. After this procedure the node number is typically still larger than the target 30 or 60, so we use the Induced Subgraph Random Walk Sampling algorithm\footnote{From {\tt github.com/Ashish7129/Graph\_Sampling}.} to subsample the graphs to the required number of nodes. 
Since the subsampling is stochastic, the choice of random seed produces different versions of the fMRI graph\footnote{At the highest threshold which required the mildest subsampling, these versions did not differ much.}. Subsequently, we transform the obtained undirected graph into a DAG using our standard procedure.

Let us note, however, that one should not treat the fMRI graphs (and especially the corresponding neural networks) as providing a realistic model of how the human brain processes information in a \emph{visual classification task}. The latter process occurs of course on a much smaller scale than the brain-wide networks. 

The interest in using the graphs based on fMRI data is that they encode some \emph{global} features of information processing by the brain.
Moreover, these graphs are not produced by a standard mathematical graph generating algorithm, so they bring an interesting variety to the range of considered networks.


\section{Key Results}
\label{s.keyresults}

In this section we first exhibit the inadequacy of classical graph invariants to select the best performing networks and describe the generic features of worst networks. 
Then we introduce a class of well performing networks (which we call \emph{quasi-1-dimensional} or Q1D) and provide their characterization in terms of a novel numerical graph invariant.

\begin{figure}[t]
\vskip 0.2in
\begin{center}
\includegraphics[width=0.48\textwidth]{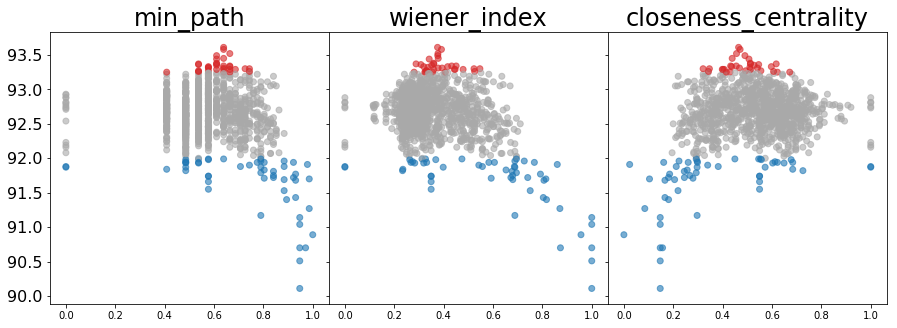}
\caption{The test accuracy versus selected network features. We indicate the best (equal or above 93.25\%) models as red, the worst (below 92\%) as blue, and the rest as gray. The presented features are able to sort-out the worst performing networks, but not the best ones.
For more details on data processing refer to \textbf{Supplementary Materials~D}.  }
\label{fig.3features}
\end{center}
\vskip -0.2in
\end{figure}

\begin{figure}[ht]
\begin{center}
\centerline{\includegraphics[width=0.4\textwidth]{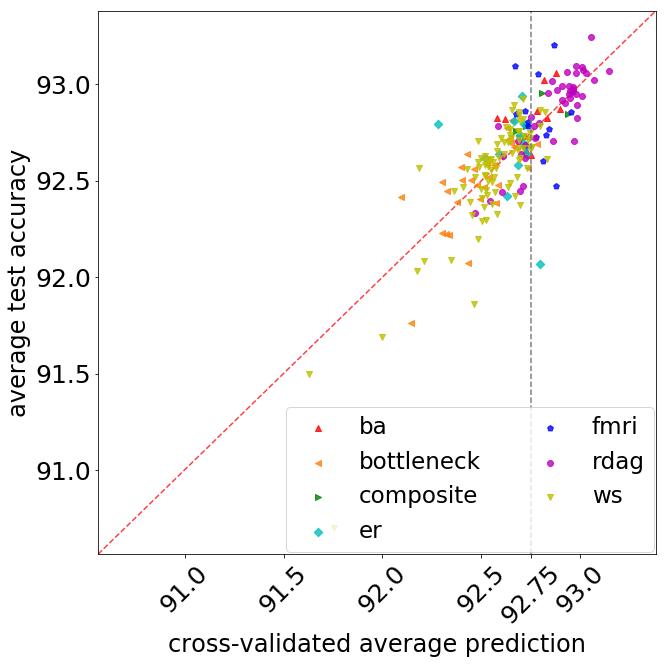}}
\caption{Average test accuracy as a function of the corresponding (cross-validated) average prediction of a Random Forest regressor with $10$ trees for the graph types with 60 nodes. The averaging is done over the graphs differing only by the random seed. It may be observed that the model is able to identify a group of best performing networks - notice the high test accuracy range for  \emph{cross-validated average prediction} $\gg 92.75$ (denoted as gray dashed line in the picture). For more details and models see \textbf{Supplementary Materials~E}.}
\label{fig.randomforest}
\end{center}
\vskip -0.4in
\end{figure}

\subsection{The Inadequacy of Classical Graph Characteristics} 
\label{s.graphinv}

\begin{figure*}[t]
\vskip 0.2in
\begin{center}
\centerline{\includegraphics[height=7cm]{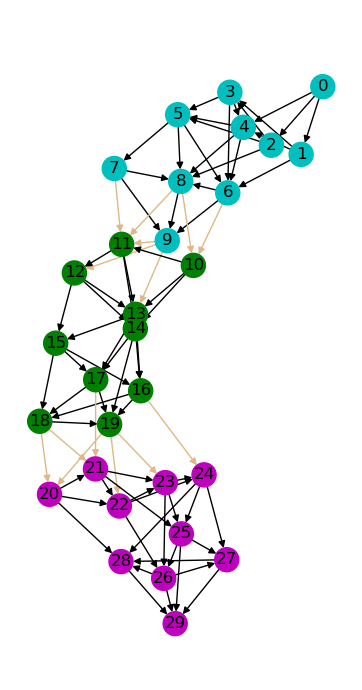}\hfill\includegraphics[height=7cm]{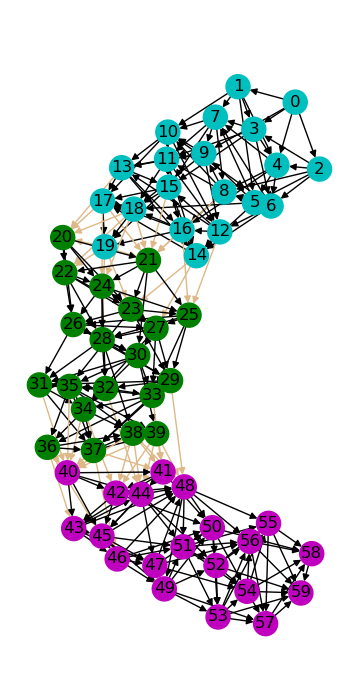}\hfill\includegraphics[height=7cm]{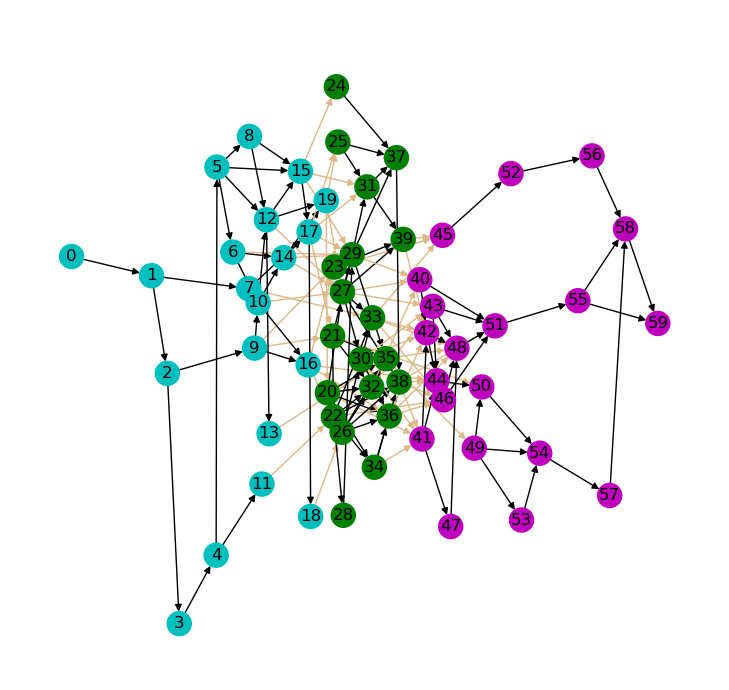}}
\caption{The best network with 30 nodes (left), with 60 nodes (center) and an example of a highly ranked fMRI based network. For more examples of the best networks see \textbf{Supplementary Materials~H}.}
\label{fig.best}
\end{center}
\vskip -0.2in
\end{figure*}

The key motivation for this work was to understand what features of the underlying graph are correlated with the test performance of the corresponding neural network. To this end, for the analysis we use $54$ graph features, mostly provided by the {\tt networkx} library as well as some simple natural ones, like the logarithm of the total number of paths between the input and output or the relative number of connections between stages with various resolutions. For a full list of the features see {\bf Supplementary Materials D}.

It turns out that none of the classical features by itself is enough to isolate the best performing networks. However, the worst networks are to a certain extent extreme and can be more or less identified (see Fig.~\ref{fig.3features} for a representative example and more plots in the {\bf Supplementary Materials~F}).

We analyzed various ML regressors (for predicting test accuracy of the given graph type) or classifiers (for predicting the best performing networks) as well as feature selection procedures. See Fig.~\ref{fig.randomforest} for the results for a Random Forest regressor. The Random Forest nicely selects a class of best performing graphs, indicating that there indeed exists a non-trivial relation between the network test accuracy and the topology of the underlying graph. However, the complexity of the model makes it difficult to interpret. Therefore, in the following, we will introduce new simple numerical characteristics which will pick out a range of well performing graphs.

\subsection{The Worst Networks} 

\begin{figure}[ht]
\begin{center}
\centerline{\includegraphics[height=4cm]{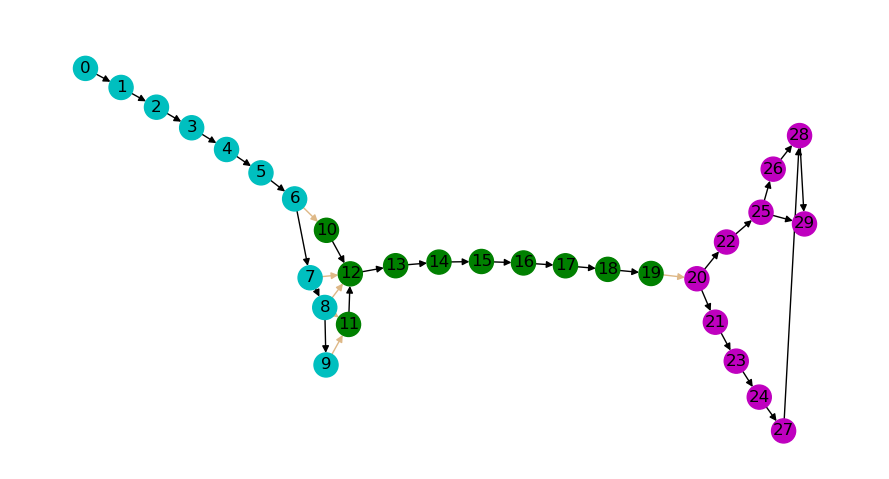}}
\caption{One of the worst networks with 30 nodes. The worst networks are typically characterized by sparse connections and long chains of operations. For more examples of the worst networks see \textbf{Supplementary Materials~H}.}
\label{fig.bad}
\end{center}
\vskip -0.2in
\end{figure}

As mentioned before, several investigated network features seem to be able to discriminate the worst networks (Fig.~\ref{fig.3features}). Those networks are usually characterized by long distances between any two nodes in the graph, resulting in long chains of operations and sparse connections. An example of such a graph is presented in Fig.~\ref{fig.bad}. In addition, we verified that purely sequential 1d chain graphs (node $i$ is connected only to node $i+1$) gave indeed the worst performance.

\begin{figure}[ht]
\begin{center}
\centerline{\includegraphics[height=4.5cm]{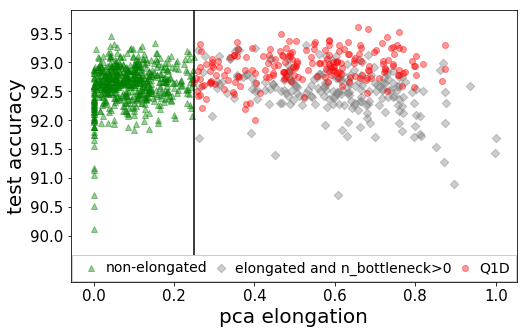}}
\caption{The visualization of the Q1D criterion. The green triangles indicate graphs without a global elongated structure and the gray diamonds are used to represented the elongated graphs with bottlenecks. Networks with Q1D property are drawn as red dots. The black vertical line illustrates the $pca\ elongation$ division point at $0.25$. The Q1D criterion successfully selects the best networks from the elongated group. }
\label{fig.q1dscatter}
\end{center}
\vskip -0.2in
\end{figure}

\subsection{The Best Networks -- quasi-1d Graphs and Others} 
\label{s.Q1D}

We observe that the best networks 
belonged predominantly to the Random DAG category with short range connections (i.e. exponential $f(x)$). One generic visual feature of these graphs is their quasi-1d structure\footnote{Here we use this term intuitively. We will provide a precise definition shortly.} (see the first two graphs in Fig.~\ref{fig.best}) - there is a definite global ordering in the feed-forward processing sequence defining the 1d structure, yet locally there are lots of interconnections which most probably implement rich expressiveness of intermediate feature representations. These models have a very large number of paths between the input and the output. This is, however, not the feature responsible for good performance, as maximally connected DAGs which have the maximal possible number of paths do not fall into this category and have worse performance (see Fig. \ref{fig.bottleneck_bars} later in the paper.). 
In contrast, filament-like, almost sequential models like some Watts-Strogatz networks (recall Fig.~\ref{fig.bad}) have in fact significantly worse performance, so sequentiality by itself also does not ensure good performance.

We would like now to characterize these graphs purely in terms of some numerical graph features without recourse to their method of construction. This is not \emph{a priori} a trivial task.
On the one hand, one has to be sensitive to the quasi-1d structure. On the other hand the filament-like almost 1d graphs are quite similar in this respect, yet they yield very bad performance. So numerical graph invariants which are positively correlated with the 1-dimensionality tend to have similar or even larger values for the very bad graphs.

A condition which eliminates the filament-like graphs is $n_{bottlenecks}=0$, where a bottleneck edge is defined by the property that cutting that edge would split the graph into two separate components.

In order to numerically encode the quasi-1d character of a network, we perform PCA on the set of node coordinates returned by the Kamada-Kawai embedding and require sufficiently anisotropic explained variance ratio.  Please note that despite appearances this is a quite complex invariant of the original abstract graph, as the Kamada-Kawai embedding depends on the whole global adjacency structure through the spring energy minimization. Hence the nature of the embedding encodes nontrivial relevant information about the structure of the graph. We define then the elongation of the network as
\eq
pca\_elongation = 2 \cdot (variance\_ratio -0.5),
\eqx
where $variance\_ratio$ is the percentage of the variance explained  by the component corresponding to the largest eigenvalue computed during the PCA decomposition. We define the quasi-1d graphs (Q1D) as satisfying the condition: 
\eq
\label{e.q1d}
pca\_elongation>0.25 \quad \text{and} \quad n_{bottlenecks}=0.
\eqx
This condition is visualized in Fig.~\ref{fig.q1dscatter}. The first term of the Q1D definition accounts for networks which have a global one-dimensional (hierarchical) order (like the two first networks in Fig.~\ref{fig.best}) and the specific cut-off value $0.25$ is a visual estimate motivated also by Fig.~\ref{fig.q1dscatter}. The second condition eliminates graphs containing bottlenecks which form the bulk of badly performing elongated graphs (denoted by gray dots in Fig.~\ref{fig.q1dscatter}). 

We find that among the top 50 networks, $68\%$ have the Q1D property. Moreover, out of the remaining 970 graphs, only $17\%$ are Q1D. A breakdown of the top-50 and bottom-50 by specific graph families and the Q1D property is presented in Table~\ref{tab.best}. One may observe that Q1D successfully selects almost every of the best preforming \rrdags and half of the \fmri graphs (fourth  column). Those two families are also the most representative among top-50. Furthermore, \emph{none} of the graphs in the bottom-50 has the Q1D property (last column).

The Q1D criterion is able to single out \emph{one type} of the best performing networks, being at the same time \emph{agnostic} about the details of the graph generation procedure. This is especially important considering the failure of classical graph features in this regard. 

Finally, let us also mention that there are some qualitatively different networks (see for example the \fmri network in Fig.~\ref{fig.best}) in the \fmri class as well as in the \ba class which achieve good performance. Those networks are often not elongated (as indicated by several green points with high test accuracy in Fig. \ref{fig.q1dscatter}) and therefore do not satisfy the Q1D criterion. It seems, however, quite difficult to identify a numerical characterization which would pick out the best networks from this category (see e.g. Fig.~\ref{fig.randomforest} for the range of random forest predictions around and below $92.75$, which have a wide range of test accuracies). Indeed, there are also some individual highly ranked \ws networks, which have rather badly performing counterparts with the same \ws generator parameters and differing only in the random seed.

\begin{table*}[t]
\caption{ For each graph family we report in percentage the number of all graphs having Q1D property, the number of graphs in top-50, the share of the given family in top-50 and the number of Q1D graphs within the ones present in top-50, followed by analogous statistics for the graphs in bottom-50. The Q1D criterion selects almost every best performing \rdag and more than half \fmri graphs (fourth column), which are the majority in top-50 (second column). None of the worst performing graph satisfies the Q1D criterion (last column). }
\label{tab.best}
 \begin{center}
 \begin{small}
 \begin{sc}
\begin{tabular}{lr|rrr|rrr}
\toprule
model &  having Q1D & in top-50  & share in&  Q1D within & in bottom-50 & share in & Q1D within  \\

{} & property &   & top-50 & top-50 &  & bottom-50 & bottom-50  \\
\midrule

ba          & 0.00 & 4.00 &   4.00 &  0.00 &  0.00 & 0.00 & 0.00  \\
bottleneck  & 0.00 & 0.67 &  2.00 & 0.00 &  6.67 &  20.00 & 0.00  \\
composite   & 0.00 & 0.00 & 0.00 &  0.00  &  0.00 &  0.00 & 0.00  \\
er          & 1.33 & 2.67  &  4.00 &  0.00  &  2.67 &  4.00 & 0.00 \\
fmri        & 32.86 & 12.86  & 18.00 & 55.56 &  1.43 &  2.00 & 0.00 \\
rdag        & 66.98 & 13.95  & 60.00 & 93.33  &  0.93 &  4.00 & 0.00 \\
ws          & 7.05 & 1.36 & 12.00 & 16.67  &  7.95 & 70.00 & 0.00  \\
\midrule
all         & 19.50 & 4.90 & -- & {\bf 68.00} & 4.90 & -- & {\bf 0.00} \\

\bottomrule
\end{tabular}
 \end{sc}
 \end{small}
 \end{center}
\end{table*}

\section{Further architectural results}
\label{s.architecture}

In this section we analyze the interrelation with neural network performance of such architectural features as the number of resolution changing pathways, the effect of short vs. long range connections and the interplay of depth and width with the Q1D property introduced in the previous section. We also perform a comparison of the CIFAR-10 results with results on CIFAR-100 in order to ascertain the consistency of the identification of the best and worst performing network families.

\subsection{Resolution Changing Pathways. The Impact of Bottlenecks} 
\label{s.bottleneck}

\begin{figure}[hbt]

\begin{center}
\centerline{\includegraphics[width=0.46\textwidth]{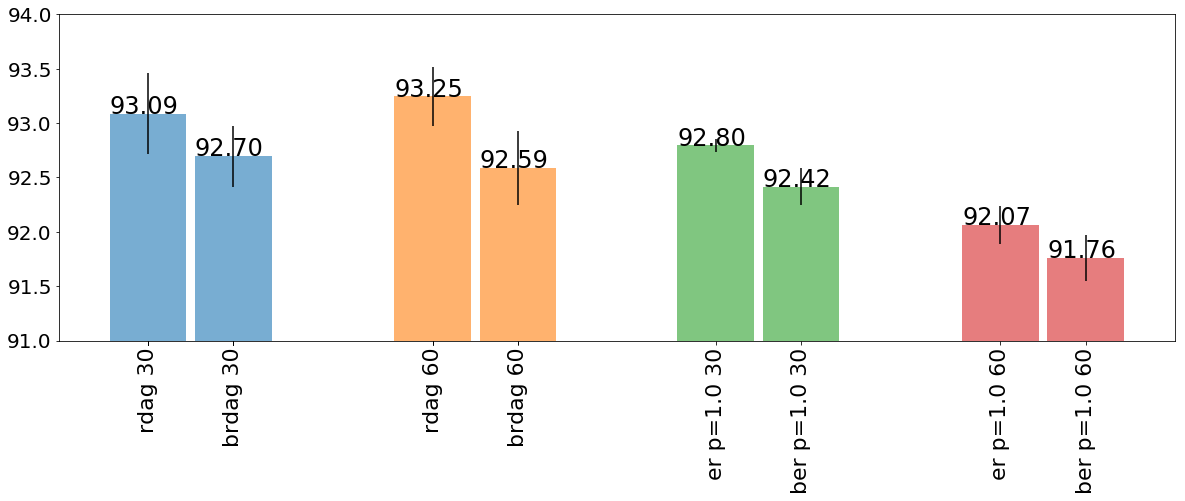}}
\caption{The CIFAR-10 test accuracy for selected pairs of networks and their bottleneck ({\tt b} prefix) ablations. From left: best \rdag with $30$ nodes, best \rdag with $60$ nodes, maximally connected DAG (\er with $p=1.0$) on $30$ and $60$ nodes. In all cases the introduction of the bottleneck leads to worse mean performance. }
\label{fig.bottleneck_bars}
\end{center}
\vskip -0.2in
\end{figure}

As noted in section~\ref{s.nndetails}, one difference between the networks of \cite{xie2019exploring} and our construction was that in the former case, there were separate random graphs for each processing stage of a specific resolution, which were connected with a single gateway. In our case we have a single graph which encompasses all resolutions. Thus generally there are many independent resolution reducing edges in the network instead of a single one.

\begin{figure}[htb]

\begin{center}
\centerline{\includegraphics[height=3.2cm]{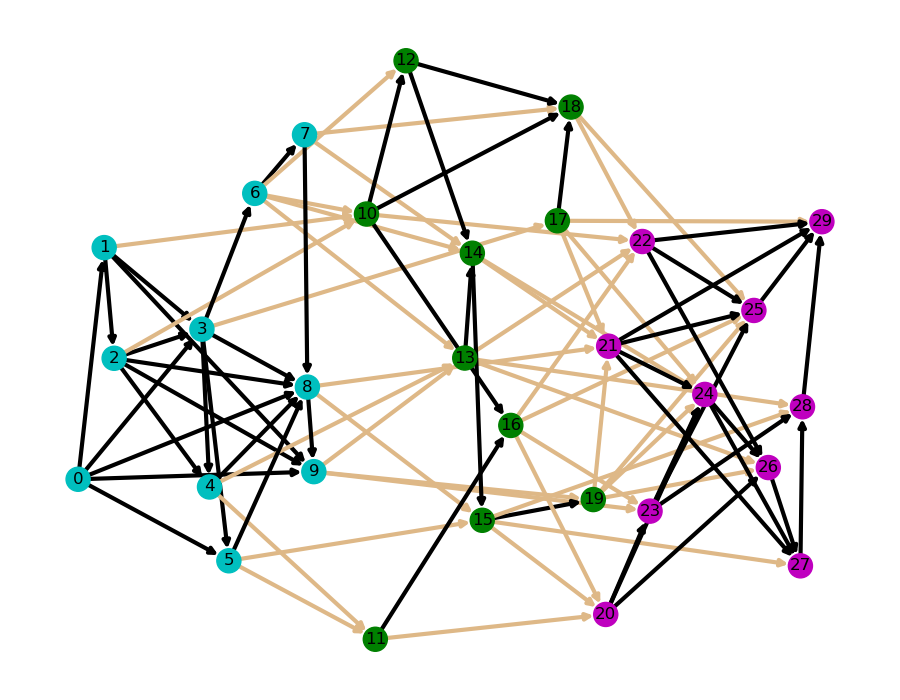}\hspace{0.2cm}\includegraphics[height=3cm]{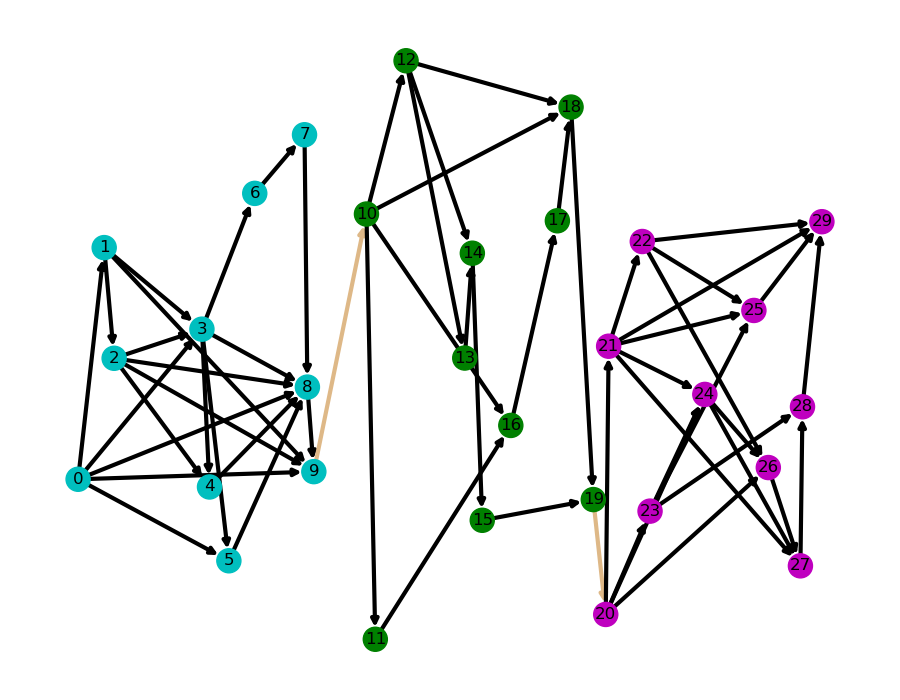}}
\caption{Original graph (left) and its bottleneck variant (right).}
\label{fig.bottleneck}
\end{center}
\vskip -0.2in
\end{figure}

In order to verify whether such a single gateway between different resolutions is beneficial or not, for a selected set of graphs we artificially introduced such a bottleneck by first erasing all inter-resolution edges. Next, we create a single edge from the last node in the preceding stage to first node in the consequent stage (this is illustrated in Fig.~\ref{fig.bottleneck}) and then fixing possible orphans as in Fig.~\ref{fig.orphan}.

We found that, systematically, the introduction of a bottleneck deteriorates performance (see Fig. \ref{fig.bottleneck_bars}). Hence multiple resolution reduction pathways are beneficial. Let us note that this result is coherent with our findings from section~\ref{s.Q1D}, where bottleneck edges (also within a single resolution stage) typically appear in badly performing networks and hence are excluded from the definition of Q1D.


\begin{figure}[h!]
\vskip 0.2in
\begin{center}
\includegraphics[width=0.48\textwidth]{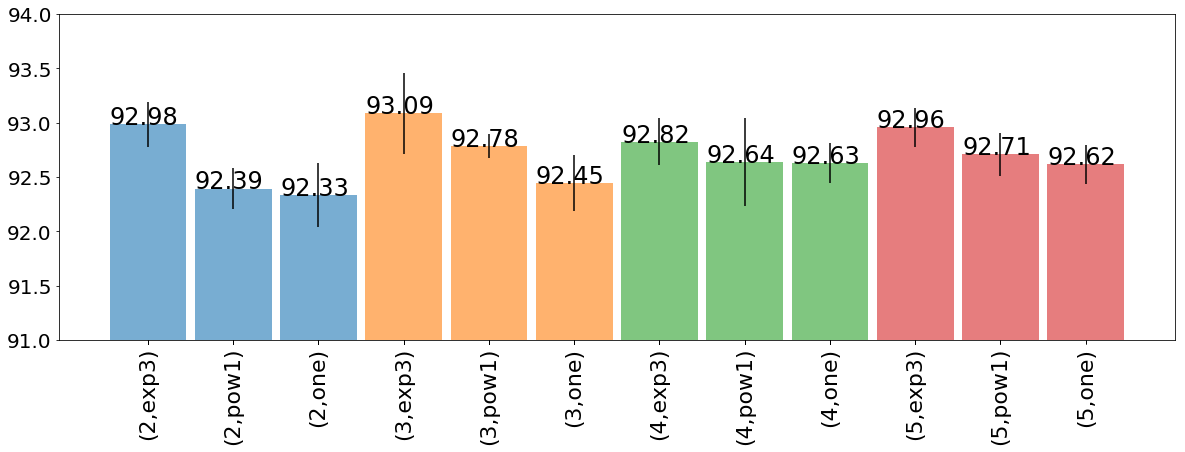}
\caption{The CIFAR-10 test accuracy averaged over different versions (random seeds) of random DAG models with $30$ nodes and constant number ($2$-$5$) of output edges $n_i^{out}$. 
The symbol \emph{exp3} stands for exponential weighting function $f(x)$,  \emph{pow1} for a power law and \emph{one} for a constant. It may be observed that the networks with primarily local connections (\emph{exp3} - the first bar in each set) have the best performance. }
\label{fig.longshortbars}
\end{center}
\vskip -0.2in
\end{figure}

\subsection{Long- vs. Short-range Connections} 

\begin{figure*}[t]
 \vskip 0.05in
    \centering
    \begin{subfigure}[t]{0.42\textwidth}
        \centering
        \includegraphics[width=\linewidth]{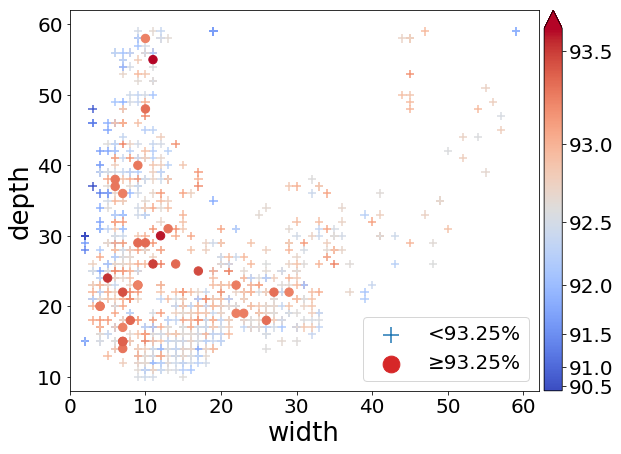}
        \caption{all networks}
        \label{fig.depthwidth_a}
    \end{subfigure}%
    ~
    \begin{subfigure}[t]{0.42\textwidth}
        \centering
        \includegraphics[width=\linewidth]{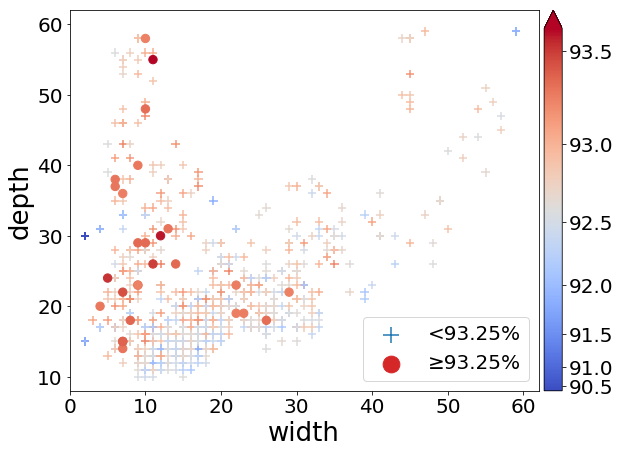}
        \caption{$n_{bottlenecks} = 0$}
        \label{fig.depthwidth_b}
    \end{subfigure}
    \\
    \begin{subfigure}[t]{0.42\textwidth}
        \centering
        \includegraphics[width=\linewidth]{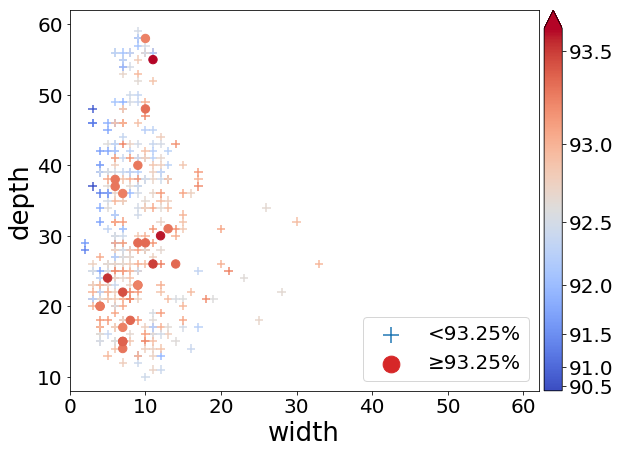}
        \caption{$pca\ elongation > 0.25$}
        \label{fig.depthwidth_c}
    \end{subfigure}
    ~
    \begin{subfigure}[t]{0.42\textwidth}
        \centering
        \includegraphics[width=\linewidth]{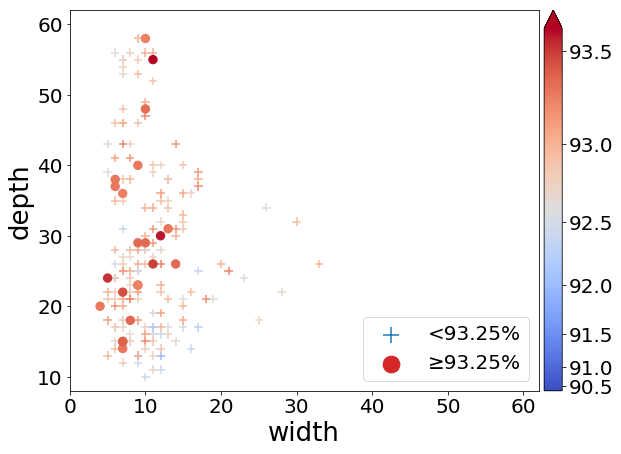}
        \caption{Q1D networks}
        \label{fig.depthwidth_d}
    \end{subfigure}
    \caption{From left: the scatter plots of the depth and width for \textbf{(a)} all the graphs, \textbf{(b)} graphs with no bottleneck edges, \textbf{(c)} elongated graphs and \textbf{(d)} graphs satisfying the Q1D criterion. The colorbar indicates the networks test accuracy. The best performing networks (accuracy greater than or equal to $93.25\%$) are represented using dots, while the remaining networks are represented using crosses. It may be observed that the depth and width alone are not able to predict the best performance (Fig. \textbf{(a)}). Although the best performing networks have typically small width, the same applies for the worst ones.  Combining the bottleneck and $pca\ elongation$ requirements makes it possible to sort out a group of best performing networks with small width (Fig \textbf{(d)}). }
    \label{fig.depthwidth}
\end{figure*}

The algorithm for directly generating random DAGs allows for modifying, in a controllable way, the pattern of long- versus short-range connectivity. This is achieved by changing the function $f(x)$ from an exponential, leading to local connections, through a power law, which allows for occasional long range connections, to a constant function, which does not impose any spatial order and allows connections at all scales. The results are presented in Fig. \ref{fig.longshortbars}. We observe that within this class of networks the best performance comes from networks with primarily short range connections and deteriorates with their increasing length.

This may at first glance seem counter-intuitive, as skip connections are typically considered as beneficial. However, the effect of long range connections which is associated with easier gradient propagation is already taken care of by the residual structure of each node in our neural networks (recall section~\ref{s.nndetails}). One can understand the deterioration of the network performance with the introduction of long term connections as coming from an inconsistency of the network with the natural hierarchical semantic structure of images. This result leads also to some caution in relation to physical intuition from critical systems where all kinds of power law properties abound. The dominance of short-range over long-range connections is also consistent with the good performance of \emph{quasi-1-dimensional} networks as discussed in section~\ref{s.Q1D}.

\subsection{Depth and Width are Not Enough}

In addition to the studies presented in the preceding sections, we investigate the performance of the networks and the intuition behind the Q1D definition from the perspective of the two most often used global network features: network depth and width. Since the networks studied in this paper do not follow the standard sequential computation pattern, we need to appropriately reformulate those characteristics. 

Given a network based on a DAG we define the \emph{depth} as the maximal length of a path from the input to the output node. The \emph{width} is defined as the maximal number of nodes that need to be maintained during the feed forward computation of the model. More formally, we say that the width $W_n$ at node $n$ is equal to the number of nodes $u$ such that $u<n$ and there exists an edge $(u,w)$ such that $w\ge n$. The width of the network is then defined as the maximum node width $\max_n W_n$ over all graph nodes. 

The interrelation of the networks depth and width with the test accuracy is presented in Fig. \ref{fig.depthwidth_a}. One may observe that depth is not predictive of a network performance. Indeed, many best networks have very distinct (even quite shallow) maximum path lengths. Similarly, although the models achieving high test accuracy have width smaller than $30$, this is not a selective feature, as most of the worst models also fall into this category. It is also worth mentioning that graphs which are both very deep and very wide (such as for example the fully-connected DAGs) perform rather poorly. 

Analyzing the same plot narrowed down to graphs with no bottleneck edges (Fig. \ref{fig.depthwidth_b}) reveals that this criterion is successful in sorting out the best networks among those rather deep and thin. On the other hand, requiring only elongated graphs (Fig. \ref{fig.depthwidth_c}) excludes the wide models, as well as the poorly performing group of networks with both small depth and width. Combining those conditions as in the definition of Q1D (\ref{e.q1d}) singles out one group of the best networks -- the \emph{quasi-1-dimensional} ones (Fig. \ref{fig.depthwidth_d}).  

To summarize, in the context of neural networks built on arbitrary graphs, we observe a surprisingly complex panorama of the interrelation of depth and width with the network performance. One cannot, therefore, restrict oneself to using just these observables as parameterizing test accuracy. A more fine-grained analysis of the graph structure is necessary, like the one done for \emph{quasi-1-dimensional} networks.


\begin{figure}[ht]
\vskip 0.2in
\centerline{
\includegraphics[width=0.35\textwidth]{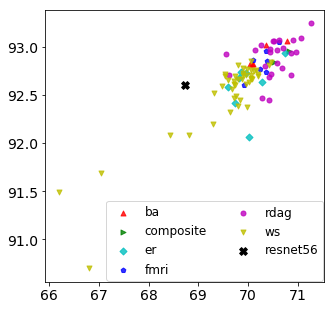}}
\caption{The CIFAR-10 (y-axis) and CIFAR-100 (x-axis) test accuracies. Each datapoint contains results averaged over the random versions of the models. The results are strongly correlated, yielding Pearson correlation coefficient equal to $0.868$.}
\label{fig.cifar10vs100}
\vskip -0.2in
\end{figure}

\subsection{CIFAR-10 versus CIFAR-100 Consistency}
\label{s.cifar100}

In addition to the CIFAR-10 task, we trained all networks with $60$ nodes (except for the bottleneck ablations) on the CIFAR-100 dataset. We used the same training procedure as the one for CIFAR-10. The motivation for this experiment was to verify whether the graph families which performed best in the first problem achieve also high results in the second. Indeed, we observe a significant correlation $0.868$ (see Fig.~\ref{fig.cifar10vs100}) between the respective test accuracies (averaged over the 5 random realizations of each graph type). 
Especially noteworthy is the consistency between the groups of best and worst graphs for the two datasets.
 

\section{Summary and Outlook}

We have performed an extensive study of the performance of artificial neural networks based on random graphs of various types, keeping the training protocol fixed. The number of parameters for each network was approximately fixed to be equal to the number of parameters of a ResNet-56 network for CIFAR-10.

Apart from using the classical families of random graphs (Erd\H{o}s-R\'enyi, Barab\'asi-Albert and Watts-Strogatz), we introduced an algorithm for directly generating random directed acyclic graphs (\rdag), which is very flexible and can be tuned to generate DAGs of various types. In particular, it is well suited for modifying short- and long-range connectivity. In addition, we constructed a family of graphs based on resting state fMRI connectivity networks of the human brain. 

One class of networks which had the best performance in our simulations (clearly better than the reference ResNet-56 model) were networks which could be characterized as quasi-1-dimensional, having mostly local connections with a definite 1-dimensional hierarchy in data processing (one can dub this structure as {\it local chaos and global order}).
These were predominantly networks in the \rdag family.
We also introduced a very compact numerical characterization of such graphs.

It is worth noting, that some of the fMRI based graphs were also among the best performing ones (together with some \ws and \ba ones). 
We lack, however, a clear cut numerical characterization of these ``good'' graphs as there exist graphs with apparently similar structure and numerical invariants
but much worse performance. 

Among other structural observations made in this paper, we noted that long range connections were predominantly negatively impacting network performance. Similarly, artificially imposing a bottleneck between the processing stages of various resolutions also caused the results to deteriorate.   

Thus a general guideline in devising neural network architectures which can be formed in consequence of our study is to prefer networks with mostly local connections composed into an overall hierarchical computational flow, with multiple resolution reducing pathways and no bottleneck edges.  These characteristics seem to lead most consistently to good performance among the vast panorama of connectivity patterns investigated in the present paper. 

We believe that the performed research will open up space for numerous further investigations. The massive dataset\footnote{The graph architectures are available at \url{https://zenodo.org/record/3700845}. The test accuracies as well as the PyTorch code is available on github: \url{https://github.com/rmldj/random-graph-nn-paper} } could be used for independent further exploration of the interrelation of graph topology and network performance.
The best and worst performing classes of networks identified here may be used to focus further research in specific directions.

We expect that once we move to a greater number of nodes, we may see much more marked differences between the various network types, as the main random graph families are really defined asymptotically and for a small number of nodes may tend to blend between themselves for some choices of parameters.
Considering larger graphs would be especially interesting in view of the flexibility of the random DAG algorithm introduced in this paper, which allows to generate a wide variety of networks, of which we studied only a subset here.

Another interesting direction of research is the modification of the precise neural network counterparts of the graph nodes. In this paper we adopted to some extent the formulation of \cite{xie2019exploring}, but there is definitely room for significant changes in this respect.

Finally, let us note that apart from any practical applications in the search for better network architectures, the results on the neural network performance as a function of graph architecture should yield a lot of data which could contribute to the theoretical quest for the understanding of the efficacy of deep learning.


\section*{Acknowledgements} This work was supported by the Foundation for Polish Science (FNP) project \emph{Bio-inspired Artificial Neural Networks} POIR.04.04.00-00-14DE/18-00.

The fMRI partial correlation matrix data were provided by the Human Connectome Project, WU-Minn Consortium (Principal Investigators: David Van Essen and Kamil Ugurbil; 1U54MH091657) funded by the 16 NIH Institutes and Centers that support the NIH Blueprint for Neuroscience Research; and by the McDonnell Center for Systems Neuroscience at Washington University.

\bibliography{paper}
\bibliographystyle{preprint20}

\section*{Supplementary Materials}
\appendix

\section{Training Regime}

All the networks are trained for 100 epochs on the CIFAR-10 dataset with the standard train-test split. All models are
optimized using the SGD algorithm with batch size 128, initial learning rate 0.1 and momentum 0.9. In addition, we use
a weight decay equal to 1e--4. 
The learning rate is decreased to 0.01 and 0.001 in the 80th and 90th epoch respectively. This setting is the same as the one used in ResNet, the only exception is that we train for less epochs and therefore perform the learning rate drop earlier.  The number of the initial channels C, which characterizes the size of the model, is set for each network separately, so that the total number of parameters is approximately the same as in the ResNet-56 model for CIFAR-10 (853k). For the experiment on CIFAR-100 in section 5, we use the same setting as above and perform the train and evaluation on the standard train and test split.

The code for the experiments was prepared in the \python programming language, with the use of \pytorch and \networkx packages. For the ResNet model we use the implementation provided by Yerlan Idelbayev in \url{https://github.com/akamaster/pytorch_resnet_cifar10}. The training of neural network models was conducted on GeFroce RTX 2080 Ti and GeForce GTX 980 graphic cards. The mean time of one epoch was $56.12$s ($\pm 26.59$).

We would have liked to also evaluate our networks on a much larger and more challenging  dataset (e.g. Image-Net). However within the computational infrastructure available to us, training one model on such data would require circa 17 days, which prohibits any wide-range study. In contrast, less difficult datasets than CIFAR-10 (for example MNIST or Fashion-MNIST) would not be challenging enough in order to notice the differences in the architecture. CIFAR-10 provides therefore a good trade-off between the difficulty of the problem and its size, being at the same time a well-known and balanced dataset used often in many comparison experiments.   

\begin{figure}[ht]
\vskip 0.2in 
\centerline{
\includegraphics[width=0.51\textwidth]{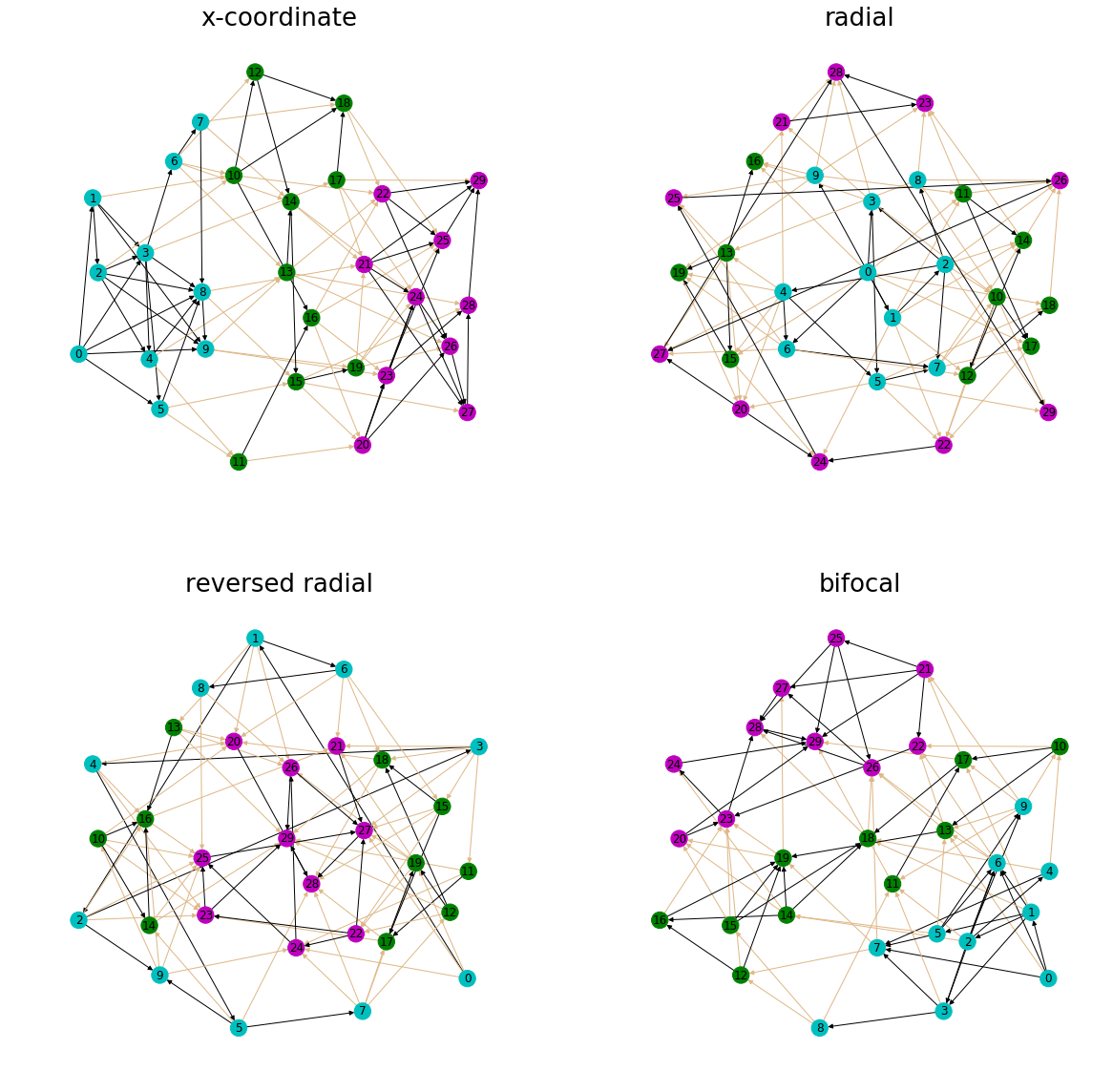}}
\caption{The visualisation of different graph orderings on the Kamada-Kawai embedding of a single graph.}
\label{fig.embedding}
\vskip -0.2in
\end{figure}

\section{From a Graph to a Directed Acyclic Graph}

\label{s.graphtodag}

\paragraph{The construction of a DAG from a graph.}
Within the framework of the present paper, where we consider standard neural networks with no recurrent connections, we need to deal with a directed acyclic graph (DAG). Yet, the vast majority of families of random graphs involve undirected graphs and so do not fall into this category. Therefore, we need a general procedure for picking the directionality of each edge. This can be achieved by picking a total ordering of the nodes of the graph (e.g. by labeling the nodes by integers) and setting the directionality of an edge from the lower to the higher node number. The key point is, of course, the precise method how to introduce the global ordering of the nodes.

The paper \cite{xie2019exploring} used either a random assignment of integers to the nodes, or a process which was tied to a specific random graph construction procedure. In contrast, we propose a general method, agnostic about the graph construction process, which automatically incorporates some knowledge about the graph structure. In particular, we would like the adopted directionality structure to be intuitively most economical and natural for the given graph.

To this end, we start with a 2D graph embedding, which is well suited for visualization. This could be e.g. Fruchterman-Reingold \cite{fruchterman1991graph} or Kamada-Kawai \cite{kamada1989algorithm} embedding. In our simulations we adopted the latter.
The advantage of this embedding is that the spatial structure (i.e. node coordinates) is coupled to the abstract internal connectivity structure of the graph so that the total spring energy associated to the edge lengths is minimal.

We fix the ordering by sorting the nodes by the $x$-coordinate of the Kamada-Kawai embedding and orient the edges accordingly. However, in order to put a neural network on a DAG, we need to have only a single input node and a single output node. Moreover, all paths on the DAG should arise from paths going all the way from the input to the output node. Hence we need to fix those orphan nodes by adding a connection from the node with the preceding number or adding a connection to the node with the succeeding number (this is illustrated in Fig.~\ref{fig.orphan} in the main text).

In the next section we make an investigation of other choices of graph embeddings and methods for imposing the directionality of the graph.  


\section{Comparison of Graph Node Orderings}
\label{s.comparison}

\paragraph{Comparison of embeddings.}
In this subsection we investigate the performance of a neural network in relation to the node ordering provided in the DAG transformation procedure. We fix the underlying undirected graph to a \er network with parameters $n=30$ and $p=0.2$. For both the Fruchterman-Reingold and Kamada-Kawai embeddings, we consider four methods for obtaining the ordering of the nodes based on their coordinates:
\begin{itemize}

    \item \textbf{x-coordinate} (\x) - The ordering is given by the x-coordinate of the nodes.

    \item \textbf{radial} (\radial) - The ordering is given by the norm of the coordinates.

    \item \textbf{reversed radial} (\radialr) - The reversed version of the radial ordering.

    \item \textbf{bifocal} (\bifocal) - First, two nodes with the largest minimum distance are selected. These nodes will be the output and input nodes of the network. The rest of the nodes are ordered by \(d_i=\frac{d_{i,1}-d_{i,2}}{d_{i,1}+d_{i,2}}\), where $d_{i,1}$, $d_{i,2}$, are the distances from node $i$ to the chosen input and output nodes. 
\end{itemize}

\begin{figure}[ht]
\centerline{
\includegraphics[width=0.45\textwidth]{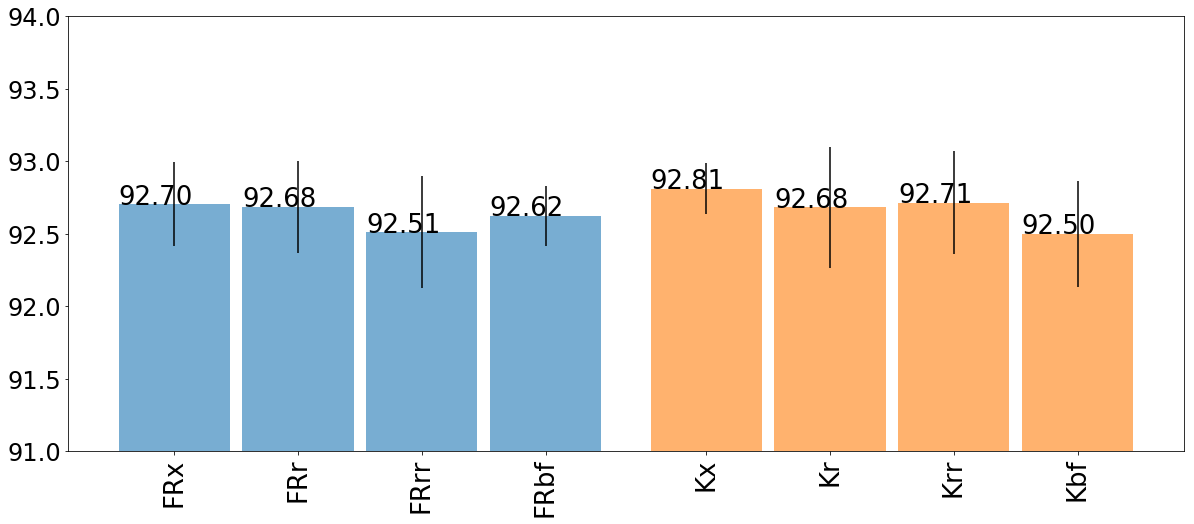}}
\caption{The test accuracy on CIFAR-10 dataset on the \er graph with 30 nodes and $p=0.2$ averaged over the versions of the model. The 'FR' and 'K' prefixes stand for the two tested graph embeddings (Fruchterman-Reingold and Kamada-Kawai). For each of the embeddings four node orderings are investigated: x-coordinate, radial, reversed radial and bifocal. The orderings are denoted in the suffix of the label respectively as \x, \radial, \radialr and \bifocal. The \x-coordinate Kamada Kawai embedding obtains the highest mean test accuracy at the same time having the lowest standard deviation.}
\label{fig.embeddings}
\vskip -0.1in
\end{figure}

The impact of those orderings on the flow of the resulting DAG is presented in Fig. \ref{fig.embedding}. The performance of the corresponding neural network models is presented in Fig. \ref{fig.embeddings}. We choose the Kamada-Kawai embedding with the x-coordinate method for the rest of the experiments, as it provides good mean results, at the same time having low standard deviation.

\paragraph{From undirected to directed DAG - node order.}

\begin{figure}[ht]
   \centering
    \includegraphics[height=4cm]{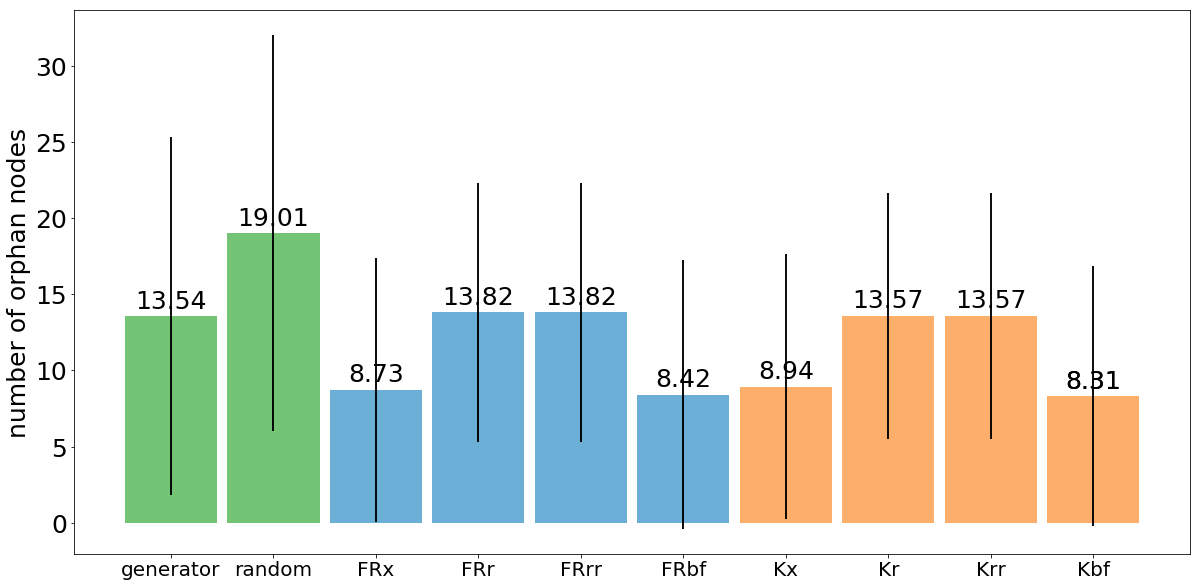}
   \caption{Number of orphan nodes for different orderings.}
   \label{fig:1aa}
   \vskip -0.2in
\end{figure}
\begin{figure}[ht]
   \centering   
    \includegraphics[height=4cm]{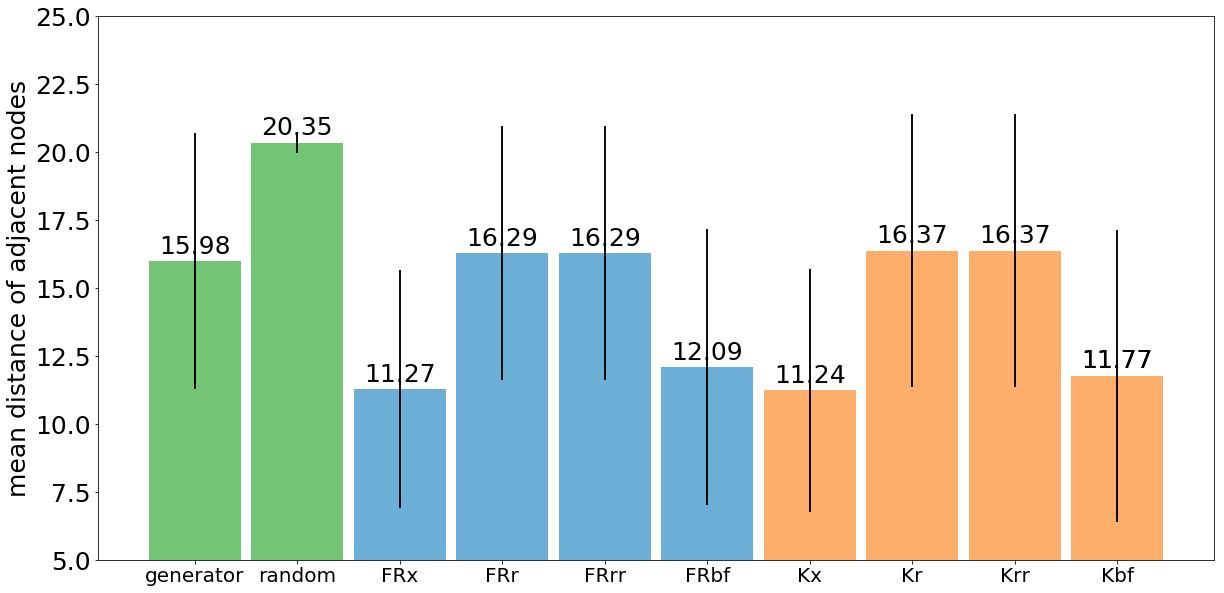}
  \caption{Mean absolute distance between indices of adjacent nodes in the graph.} \label{fig:1bb}
\vskip -0.2in
\end{figure}

In order to investigate which type of node ordering is mostly natural for a given graph, for each of the graph families and a set of generators parameters (summarized in Table~\ref{tab.networks_params_orphans}) we sample $50$ networks. We then transform them to DAGs using either the ordering returned by the generator, a random relabeling, or the embedding-based approaches from the previous paragraph. We report the mean number of orphan nodes that need to be fixed, as well as the mean absolute distance between the indices of adjacent nodes in Fig. \ref{fig:1aa} and Fig. \ref{fig:1bb} (the lower the better). One may observe that the $x$-coordinate and bifocal embedding approaches achieve the best results, indicating that these orderings require least interference after the erasing procedure and thus are more informative about the original undirected graph. 

\section{Collecting and Processing Graph Features}
\label{s.data}
For each DAG network we computed $54$ features, summarized in Table~\ref{tab.features_deff}. All attributes are rescaled using the min-max scaling. Selected features are also raised either to the power $1/2$ or $1/4$.

\section{Evaluating Various Regression Models}

We divide the dataset of graph attributes for networks with 60 nodes into train and test sets, having respectively 300 and 245 samples. Note that we do not allow for separate versions of the same graph model to appear both in test and train dataset. As the dependant variable we select the test accuracy on CIFAR-10 dataset averaged over different versions of the model. We investigate six different regression methods: Support Vector Regression (SVR), K-Nearest Neighbors (KNN), Random Forest Regressor (RF), AdaBoost (AB), Gradient Boosting Regressor (GBR) and Linear Regression (LR) with ridge, lasso and combined (Elastic Net model) regularization.\footnote{We use the implementations provided in \slearn package.} For each of them we select the best hyper-parameters by performing a 5-fold cross-validation and picking the model with the lowest mean squared error averaged over the folds. Next, we evaluate the models on the test dataset. The results are presented in Table \ref{tab.regression_models}.  For the list of considered hyper-parameters please refer to Table \ref{tab.regression_hyperparams}.

\begin{table}[ht]
\caption{The MSE and explained variance  on the test set for networks with $60$ nodes. The Random Forest (RF) achieved the best results.}
\label{tab.regression_models}
\vskip -0.15in
\begin{center}
\begin{small}
\begin{sc}
\begin{tabular}{lcc}
\toprule
 Model & MSE & Explained Variance \\
 \midrule
 svr & 0.0508 & 0.5421 \\
 knn & 0.0727 & 0.2725 \\
 rf &  \textbf{0.0245} & \textbf{0.7472} \\
 ab & 0.0344 &        0.6465 \\
 gbr &  0.0317 & 0.6732\\ 
 lr - lasso & 0.0392 & 0.6131\\
 lr - ridge & 0.0594 & 0.5006 \\
 lr - elastic-net & 0.0366 & 0.6502 \\

\bottomrule
\end{tabular}
\end{sc}
\end{small}
\end{center}
\vskip -0.2in
\end{table}

\section{CIFAR-10 Test Accuracy versus Graph Attributes}
\begin{figure*}[p]
\vskip 0.2in 
\centerline{
\includegraphics[width=0.90\textwidth]{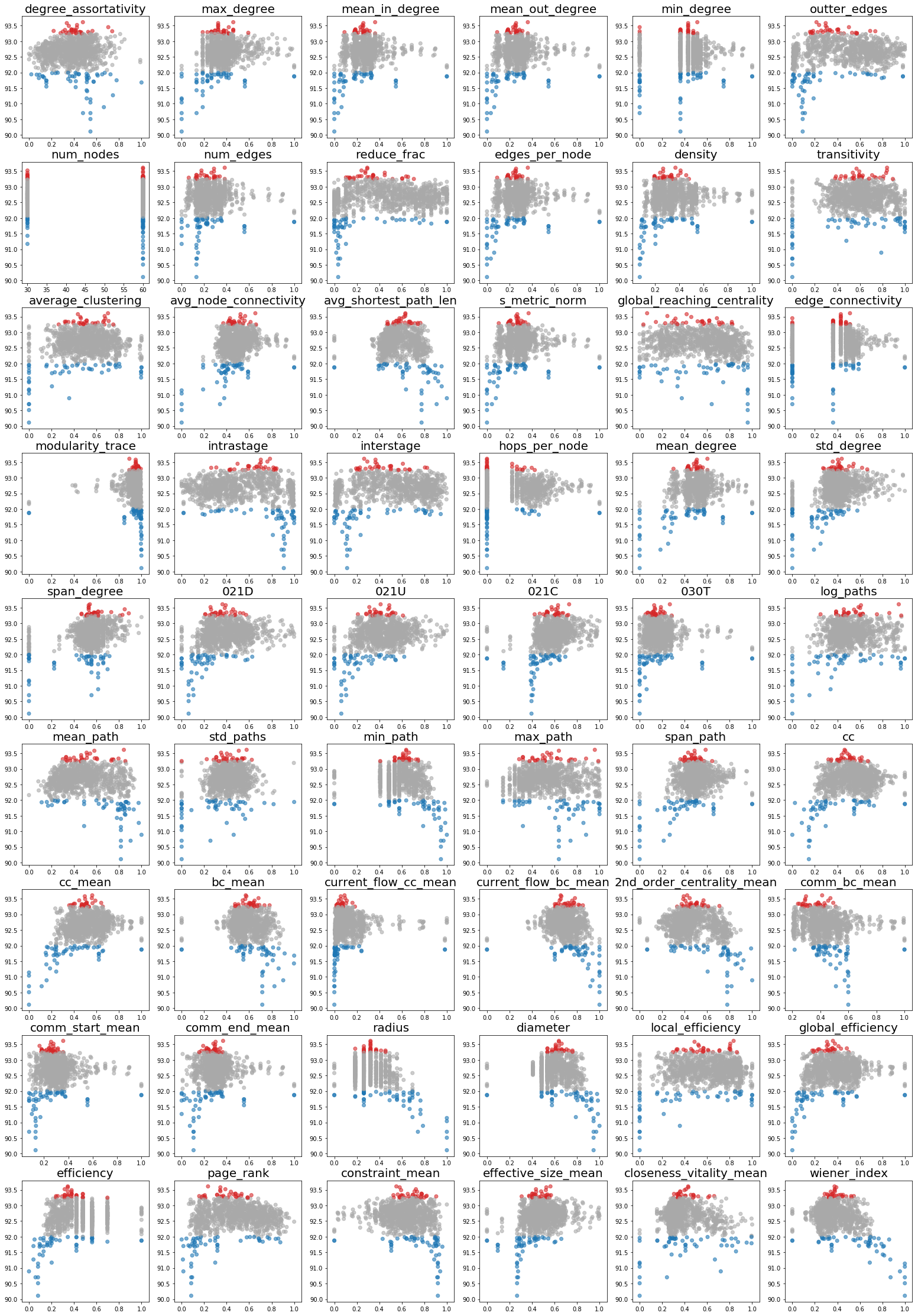}}
\caption{The test accuracy on CIFAR-10 dataset versus the investigated graph attributes.  We indicate the best (equal or above 93.25\%) models as red, the worst (below 92\%) as blue, and the rest as gray. The shortcuts 'cc', 'bc' and 'comm' stand for 'closeness centrality', 'betweenness centrality' and 'communicability'.}
\label{fig.all}
\vskip -0.2in
\end{figure*}

For every investigated graph attribute we plot the test accuracy on CIFAR-10 dataset versus the value of the attribute (after the processing described in section \ref{s.data}). The results are presented in Fig. \ref{fig.all}.

\section{Summary of the Investigated Graph Families}

In total, we have analyzed $1020$ graphs: 475 with $30$ nodes and 545 with $60$ nodes. The number of models in each graph family, as well as the investigated graph generator parameters are given in Table~\ref{tab.networks_params}. Graph generators parameters studied in the CIFAR-100 problem are summarized in Table~\ref{tab.networks_params100}. The graphs illustrated in the figures in the main text are:\\ 
\begin{footnotesize}
Fig. \ref{fig.bad}: {\tt wskx\_40\_30\_2\_v1}\\
Fig. \ref{fig.best} (from left): 
{\tt rdag\_constant\_3\_exp3\_v1}, \\
{\tt rdag\_constant\_5\_60\_v3}, {\tt fmri\_50\_60\_v0}\\
The exact methods and parameters used to generate them are: \\
method: Watts-Strogatz, nodes: $30$, $p=0.4$, $k=2$, $seed=555$.\\ 
method: Random DAG, nodes: $30$, $n_i^{out}=3$, $f(x)=\exp(-3x)$, $seed=555$.\\ 
method: Random DAG, nodes: $60$, $n_i^{out}=5$, $f(x)=\exp(-2x)$, $seed=331$. \\ 
method: fMRI DAG, nodes: $60$, $threshold=0.5$, $seed=1621$.
\end{footnotesize}

\section{Examples of the Best and Worst Graphs}
\label{sec:all_examples}
\begin{figure*}[ht]
    \centering
    \includegraphics[width=\textwidth]{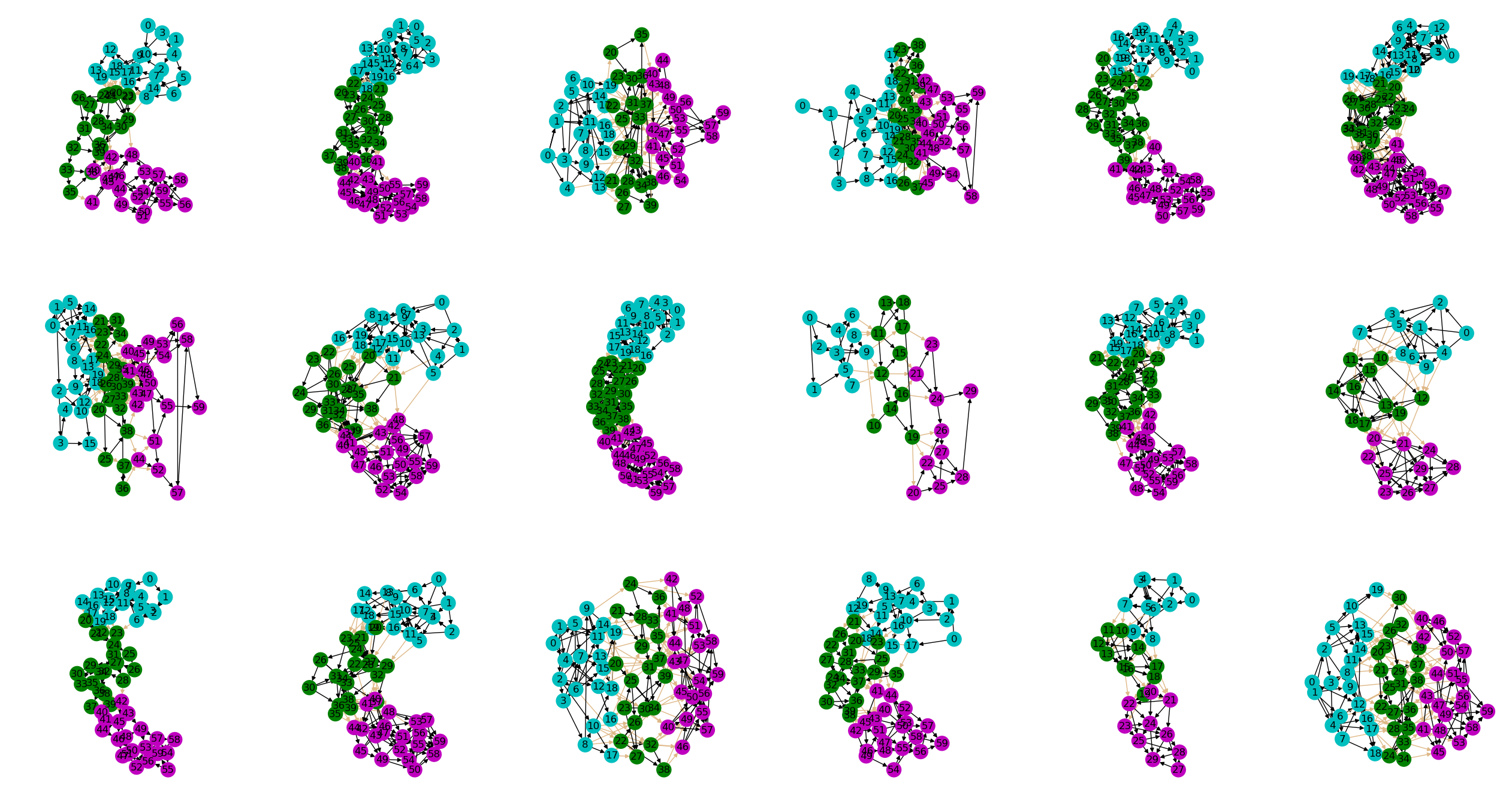}
    \caption{A selection of graphs from the top-25. The graphs are typically elongated and do not posses any bottleneck edges.}
    \label{fig:all_top}
\end{figure*}

\begin{figure*}[h]
    \centering
    \includegraphics[width=\textwidth]{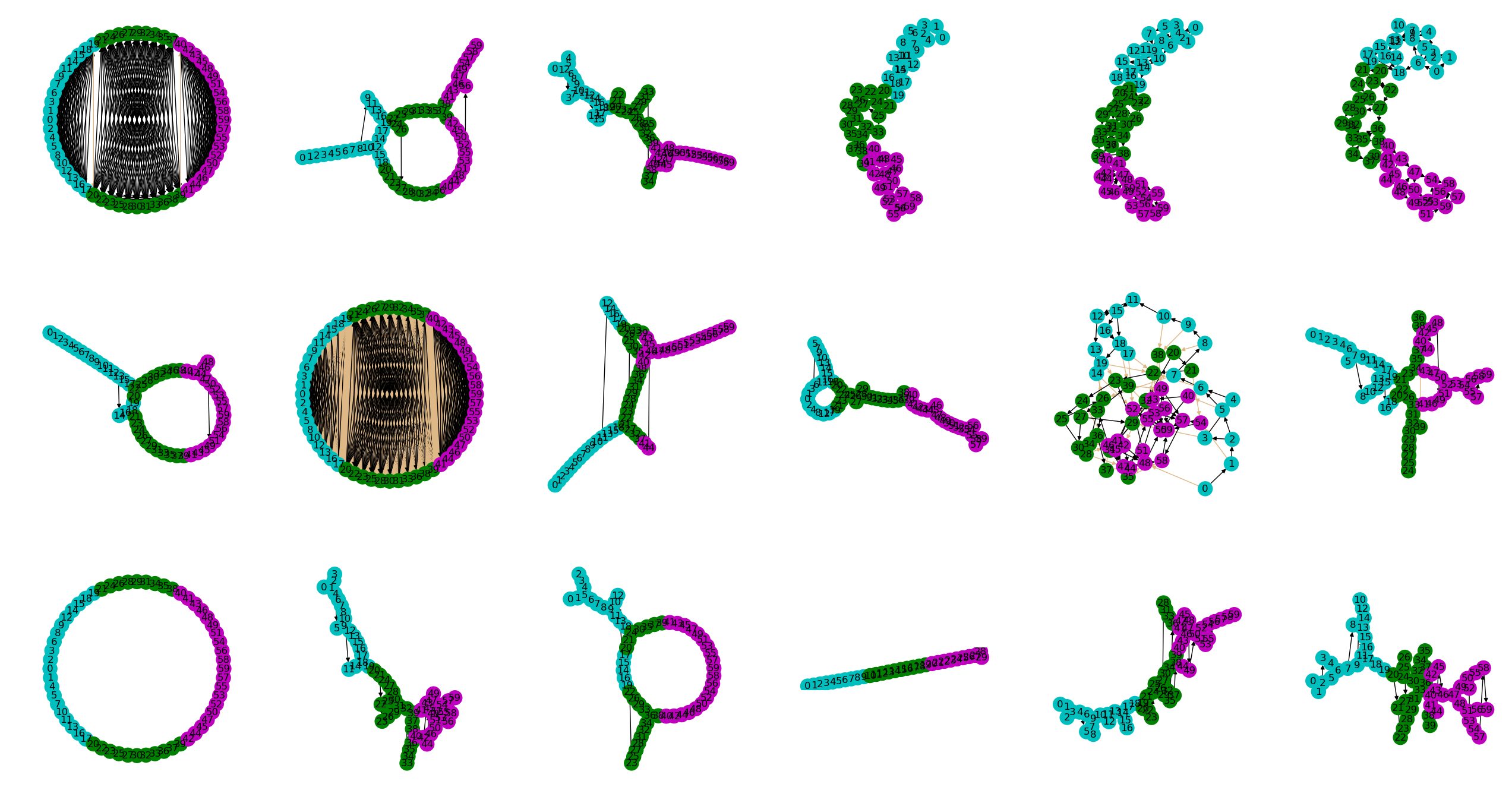}
       
    \caption{A selection of graphs from the bottom-25. The graphs are usually sparse with long chains of operations. Please note that the last three elongated graphs in the first row contain bottlenecks (and therefore are not characterized by the quasi-1-dimensional  structure). In addition, it may be observed that fully-connected graphs also do not perform well.  }
    \label{fig:all_bottom}
\end{figure*}

In Fig. \ref{fig:all_top} and in Fig. \ref{fig:all_bottom} we present a selection of graphs with performance in the top-25 and in the bottom-25, respectively. It may be observed that the best performing graphs are typically elongated and have rich local connectivity. The worst graphs are, in contrast, much more sparse and contain long chains of operations. 

\begin{table}[hbt]
\caption{The considered hyper-parameters for each model. 'DT' in AdaBoost states for Decision Tree. For the parameters names we use the same convention as the \slearn package.  }
\label{tab.regression_hyperparams}
\vskip 0.15in
\begin{center}
\begin{small}
\begin{sc}
\begin{tabular}{lll}
\toprule
model & parameter & values \\
\midrule
svr & \tt{C} & [0.001, 0.01, 0.1, \\
&\  & 1.0, 10.0, 100.0] \\
 & \tt{gamma} & [\normalfont{'scale'}, \normalfont{'auto'}] \\
\midrule
knn & \tt{n\_neighbors} & [3, 5, 7, 11] \\ 
 & \tt{weight} & \normalfont{['uniform','distance']} \\ 
  & \tt{leaf\_size} & [5, 15, 30, 45, 60] \\
  & \tt{p} & [1, 2] \\
\midrule
rf & \tt{n\_estimators} & [10, 50, 100] \\
& \tt{max\_depth} & [1, 3, 5, 7, 15, 30] \\
\midrule
ab & \tt{base\_estimator} & [DT, SVR, KNN] \\
& \tt{learning\_rate} & [0.001, 0.01, 0.1, 1.0] \\
& \tt{loss} & \normalfont{['linear', 'square',} \\
& \ &\normalfont{'exponential']} \\
\midrule
gbr & \tt{n\_estimators} & [10, 50, 100] \\
& \tt{learning\_rate} & [0.1, 0.01, 0.001] \\
& \tt{loss} &  [\normalfont{'ls'}, \normalfont{'huber'}] \\
& \tt{max\_depth} &  [1, 3, 5, 7, 15, 30]\\
\midrule
lr-ridge & \tt{alpha} & [0.001, 0.01, 0.1, \\
& \  & 1.0, 10, 100]\\
\midrule
lr-lasso & \tt{alpha} & [0.001, 0.01, 0.1, \\
& \  & 1.0, 10, 100]\\
\midrule
lr-elastic-net & \tt{alpha} & [0.001, 0.01, 0.1, \\
& \  & 1.0, 10, 100]\\
\ & \tt{l1\_ratio} & [0.25, 0.5, 0.75] \\
\bottomrule
\end{tabular}
\end{sc}
\end{small}
\end{center}
\vskip 3.8in
\end{table}

\begin{table*}[p]
\caption{The list of computed graph features together with their definitions. The last column indicates whether an additional transformation (raising to the power $1/2$ or $1/4$) was applied after the rescaling. An asterisk (*) after the feature name indicates that it was computed with the \networkx package (see \url{https://networkx.github.io/documentation/stable/reference/}for documentation). }
\label{tab.features_deff}
\vskip 0.15in
\begin{center}
\begin{small}
\begin{sc}
\begin{tabular}{lll}
\toprule
Feature Name &  Definition & P \\
\midrule
 1. degree assortativity* & \normalfont{assortativity of the graph by degree} & --  \\
2. max degree & \normalfont{maximum degree in the graph} & 1/2 \\
3. mean in degree & \normalfont{mean indegree} & 1/4 \\
4. mean out degree & \normalfont{mean outdegree} & 1/2 \\
5. min degree & \normalfont{minimum degree} & 1/4 \\
6. outter edges & \normalfont{relative number of the edges connecting different stages} & -- \\
7. num nodes & \normalfont{number of nodes} & -- \\
8. num edges &  \normalfont{number of edges} & 1/2 \\
9. reduce frac & \normalfont{number of reduce edges divided by all nodes} & -- \\
10. edges per node & \normalfont{number of edges divided by the number of nodes} & 1/2 \\
11. density* & \normalfont{density of the graph} & 1/2 \\
12. transitivity* & \normalfont{number of triangles divided by all triads} & 1/2 \\
13. average clustering* & \normalfont{average clustering coefficient of the nodes} & 1/2 \\
14. average node connectivity* & \normalfont{average local node connectivity} & 1/4 \\ 
15. average shortest path length & \normalfont{average length of all pairs of shortest paths} & 1/4 \\
16. s metric norm* & \normalfont{normalized sum of the product of nodes degrees for each edge} & 1/2 \\
17. global reaching centrality* & \normalfont{global reaching centrality of the graph} & 1/2 \\
18. edge connectivity*  & \normalfont{local edge connectivity between the input and output node} & 1/4\\
19. modularity trace*  & \normalfont{sum of eigenvalues of the modularity spectrum} & 1/2 \\
20. intrastage & \normalfont{relative number of edges within stages} & -- \\
21. interstage & \normalfont{relative number of edges crossing one stage} & -- \\
22. hops per node & \normalfont{relative number of edges crossing more than one stage} & 1/4 \\
23. mean degree & \normalfont{mean node degree} & 1/4 \\
24. std degree & \normalfont{standard deviation of the node degree} & 1/2 \\
25. span degree & \normalfont{maximum degree devided by minimum degree} & 1/4 \\
26. 021D* & \normalfont{number of 021D tradis (computed by triadic census)} & 1/2 \\
27. 021U* & \normalfont{number of 021U triads (computed by triadic census)} & 1/2 \\
28. 021C* & \normalfont{number of 021C triads (computed by triadic census)} & 1/2 \\
29. 030T* & \normalfont{number of 030T triads (computed by triadic census)} & 1/4 \\
30. log paths & \normalfont{logarithm of the number of all paths from input to output} & 1/2 \\
31. mean path & \normalfont{mean path length from input to output} & 1/2 \\
32. std paths & \normalfont{standard deviation of path lengths from input to output} & -- \\
33. min path & \normalfont{length of the shortest path from input to output} & 1/4 \\
34. max path & \normalfont{length of the longest path from input to output} & 1/2 \\
35. span path & \normalfont{max path divided by min path} & 1/4 \\
36. closeness centrality* & \normalfont{the reciprocal of the average shortest path distance to the output node} & 1/2\\
37. closeness centrality mean* & \normalfont{mean of the closeness centrality for every edge} & 1/2 \\
38. betweenness centrality mean* & \normalfont{mean of the relative number of all paths passing through given node}  & 1/4 \\
39. current flow closeness centrality mean* & \normalfont{mean of the electrical-current model for closeness centrality} & -- \\
40. current flow betweenness centrality mean* & \normalfont{mean of the electrical-current model for betweenness centrality} & 1/4 \\
41. second order centrality mean* & \normalfont{average second order centrality for each node} & 1/2 \\
42. communicability betweenness centrality mean*  & \normalfont{avergae  centrality based on communicability betweenness} & 1/2 \\
43. communicability start mean*  & \normalfont{mean communicability of the input node} & 1/2 \\
44. communicability end mean*  & \normalfont{mean communicability of the output node} & 1/2 \\
45. radius*  & \normalfont{radius of the graph} & 1/2 \\
46. diameter*  & \normalfont{diameter of the graph} & 1/4 \\
47. local efficiency*  & \normalfont{local efficiency of the graph} & 1/2 \\
48. global efficiency*  & \normalfont{global efficiency of the graph} & -- \\
49. efficiency*  & \normalfont{efficiency computed between the input and output nodes} & 1/2 \\
50. page rank*  & \normalfont{page rank of the output node} & -- \\
51. constraint mean*  & \normalfont{average constraint of the nodes} & 1/2 \\ 
52. effective size mean*  & \normalfont{average effective size of the nodes} & 1/2 \\
53. closeness vitality mean*  & \normalfont{average closeness vitality of the nodes} & -- \\
54. wiener index*  & \normalfont{the normalized wiener index (sum of all pairs shortest distances)} & 1/2 \\

\bottomrule
\end{tabular}
\end{sc}
\end{small}
\end{center}
\vskip -0.1in
\end{table*}

\newpage
\begin{table*}[t!]
\caption{The summary of analyzed graph families and their parameters for the node ordering experiment.}
\label{tab.networks_params_orphans}
\vskip 0.15in
\begin{footnotesize}
\begin{sc}
\begin{center}
\begin{tabular}{lll}
\toprule
family & total & generator parameters\\
\midrule

\ba & 250 & \normalfont{For $n=60$: number of initially connected nodes $m \in \{1,2,3,5,7\}$.}\\
\midrule
\er & 300 & \normalfont{For $n=60$: probability $p \in \{0.10, 0.15, 0.2, 0.4, 0.6, 0.8\}$}. \\
\midrule
\ws & 600 & \normalfont{For $n=60$: number of connections to the nearest neighbors in the initial ring $k \in \{2,4,6,8\}$ }\\
&\ &\normalfont{for every rewiring probability $p \in \{0.25, 0.50, 0.75\}.$}\\
\midrule
\textbf{SUM} & 1150 & \\

\bottomrule
\end{tabular}
\end{center}
\end{sc}
\end{footnotesize}
\end{table*}

\begin{table*}[t!]
\caption{The summary of analyzed graph families and their parameters for CIFAR-10.}
\label{tab.networks_params}
\vskip 0.15in
\begin{footnotesize}
\begin{sc}
\begin{tabular}{lll}
\toprule
family & total & generator parameters\\
\midrule
\ba & 50 & \normalfont{For $n=30$ and $n=60$: number of initially connected nodes $m \in \{2,3,5,7,11\}$.}\\

\midrule
\bottleneck & 150 & \normalfont{For $n=30$: ablations on \rdag with $n_{out}\in\{2,3,4,5\}$ and $f(x) \in \{\exp(-2x), \exp(-3x)\}$.}  \\ 
  &  & \normalfont{For $n=60$: ablations on \rdag with $n_{out}\in\{2,3,4,5\}$, $f(x) \in \{\exp(-2x), \exp(-3x)\}$ and $B\in\{5, 10\}$.} \\
  & & \normalfont{Both for $n=30$ and $n=60$: all \composite and all full DAGs (\er with $p=1.0$)} \\
\midrule

\composite & 20 & \normalfont{For $n=30$ and $n=60$: probability $p$ of the initial \er graph: $p \in \{0.85, 0.99$\}.} \\
\midrule

\er & 75 & \normalfont{For $n=30$: probability $p \in \{0.1, 0.2, 0.3, 0.4, 0.6, 0.8, 1.0\}$}.\\
 &  & \normalfont{For $n=60$: probability $p \in \{0.05, 0.1, 0.2, 0.3, 0.4, 0.6, 0.8, 1.0\}$}. \\

\midrule

\fmri & 70 & \normalfont{For $n=30$: connectivity threshold $t \in \{2.0, 2.5, 3.0, 3.5, 4.0, 4.5, 4.9\}$}.\\
 &  & \normalfont{For $n=60$: connectivity threshold $t \in \{2.0, 2.5, 3.0, 3.5, 4.0, 4.5, 5.0\}$ }.\\
\midrule

\rdag & 215 & \normalfont{For $n=30$: $f(x)\in\{\exp(-2x),\exp(-3x),1/x,1\}$, for every constant $n_{out} \in \{2,3,4,5\}$, $B=5$ and $\alpha=0.5$.}\\ 
&\ &\normalfont{In addition, one graph with $n_{out}$ sampled from Laplace distribution with scale $3$ and two hub graphs.} \\
&\ &\normalfont{For $n=60$: $f(x)\in\{\exp(-2x),\exp(-3x), 1/x\}$ for every constant $n_{out} \in \{2,3,4,5\}$, $B \in \{5,10\}$ and $\alpha=0.5$. }\\
\midrule
\ws & 440 & \normalfont{For $n=30$ and $n=60$: number of connections to the nearest neighbors in the initial ring $k \in \{2,4,6,8\}$ }\\
&\ &\normalfont{for every rewiring probability $p \in \{0.0, 0.1, 0.2, 0.25, 0.3, 0.4, 0.5, 0.6, 0.75, 0.8, 1.0\}.$}\\
\midrule
\textbf{SUM} & 1020 & \\

\bottomrule
\end{tabular}
\end{sc}
\end{footnotesize}
\end{table*}

\begin{table*}[t!]
\caption{The summary of analyzed graph families and their parameters for CIFAR-100.}
\label{tab.networks_params100}
\vskip 0.15in
\begin{footnotesize}
\begin{sc}
\begin{tabular}{lll}
\toprule
family & total & generator parameters\\
\midrule
\ba & 25 & \normalfont{For $n=60$: number of initially connected nodes $m \in \{2,3,5,7,11\}$.}\\

\midrule
\bottleneck & 0 &  \\
\midrule

\composite & 10 & \normalfont{For $n=60$: probability $p$ of the initial \er graph: $p \in \{0.85, 0.99$\}.} \\
\midrule

\er & 40 & \normalfont{For $n=60$: probability $p \in \{0.05, 0.1, 0.2, 0.3, 0.4, 0.6, 0.8, 1.0\}$}. \\

\midrule

\fmri & 35  & \normalfont{For $n=60$: connectivity threshold $t \in \{2.0, 2.5, 3.0, 3.5, 4.0, 4.5, 5.0\}$ }.\\
\midrule

\rdag & 120 & \normalfont{For $n=60$: $f(x)\in\{\exp(-2x),\exp(-3x), 1/x\}$ for every constant $n_{out} \in \{2,3,4,5\}$, $B \in \{5,10\}$ and $\alpha=0.5$. }\\
\midrule
\ws & 220 & \normalfont{For $n=60$: number of connections to the nearest neighbors in the initial ring $k \in \{2,4,6,8\}$ }\\
&\ &\normalfont{for every rewiring probability $p \in \{0.0, 0.1, 0.2, 0.25, 0.3, 0.4, 0.5, 0.6, 0.75, 0.8, 1.0\}.$}\\
\midrule
\textbf{SUM} & 450 & \\

\bottomrule
\end{tabular}
\end{sc}
\end{footnotesize}
\vskip 1in
\end{table*}

\end{document}